\documentclass[a4paper,fleqn]{cas-sc}

\usepackage[numbers]{natbib}
\usepackage{caption}
\usepackage{hyperref}
\usepackage{amsmath}
\usepackage{amsfonts}
\usepackage{amssymb} 
\usepackage{float}
\usepackage{makecell}
\usepackage{longtable}
\usepackage{booktabs}
\usepackage{graphicx}

\def\tsc#1{\csdef{#1}{\textsc{\lowercase{#1}}\xspace}}
\tsc{WGM}
\tsc{QE}

\begin{document}
\let\WriteBookmarks\relax
\def\floatpagepagefraction{1}
\def\textpagefraction{.001}

\shorttitle{A general reduced-order neural operator for spatio-temporal predictive learning on complex spatial domains} 

\shortauthors{Meng et al.}  

\title [mode = title]{A general reduced-order neural operator for spatio-temporal predictive learning on complex spatial domains}  

\author[a]{Qinglu Meng}





\credit{Conceptualization, Methodology, Software, Validation, Writing - Original draft}

\author[a]{Yingguang Li}[orcid = 0000-0003-4425-8073]
\cormark[1] 
\credit{Investigation, Methodology, Resources, Supervision, Writing - Review \& Editing}

\author[a]{Zhiliang Deng}
\credit{Data Curation, Methodology, Software, Validation}

\author[b]{Xu Liu}
\credit{Conceptualization, Supervision, Writing - Review \& Editing}

\author[a]{Gengxiang Chen}
\credit{Investigation, Methodology, Software}

\author[a]{Qiutong Wu}
\credit{Validation, Visualization}

\author[a]{Changqing Liu}
\credit{Investigation,  Writing - Review \& Editing}

\author[a]{Xiaozhong Hao}
\credit{Investigation,  Writing - Review \& Editing}

\affiliation[a]{organization={College of Mechanical \& Electrical Engineering, Nanjing University of Aeronautics and Astronautics, 210016, Nanjing}, country={China}}
    
\affiliation[b]{organization={School of Mechanical and Power Engineering, Nanjing Tech University, 211816, Nanjing}, country={China}}


\nonumnote{The source code of this research is available at github: \href{https://github.com/qingluM/RO-NORM}{https://github.com/qingluM/RO-NORM}.}

\begin{abstract}
Predictive learning for spatio-temporal processes (PL-STP) on complex spatial domains plays a critical role in various scientific and engineering fields, with its essence being the construction of operators between infinite-dimensional function spaces. This paper focuses on the unequal-domain mappings in PL-STP and categorising them into increase-domain and decrease-domain mapping. Recent advances in deep learning have revealed the great potential of neural operators (NOs) to learn operators directly from observational data. However, existing NOs require input space and output space to be the same domain, which pose challenges in ensuring predictive accuracy and stability for unequal-domain mappings. To this end, this study presents a general reduced-order neural operator named Reduced-Order Neural Operator on Riemannian Manifolds (RO-NORM), which consists of two parts: the unequal-domain encoder/decoder and the same-domain approximator. Motivated by the variable separation in classical modal decomposition, the unequal-domain encoder/decoder uses the pre-computed bases to reformulate the spatio-temporal function as a sum of products between spatial (or temporal) bases and corresponding temporally (or spatially) distributed weight functions, thus the original unequal-domain mapping can be converted into a same-domain mapping. Consequently, the same-domain approximator NORM is applied to model the transformed mapping. The performance of our proposed method has been evaluated on six benchmark cases, including parametric PDEs, engineering and biomedical applications, and compared with four baseline algorithms: DeepONet, POD-DeepONet, PCA-Net, and vanilla NORM. The experimental results demonstrate the superiority of RO-NORM in prediction accuracy and training efficiency for PL-STP.
\end{abstract}


\maketitle

\section{Introduction}
Spatio-temporal processes are processes that evolve in space and time \cite{harvill2010spatio}. Predictive learning for spatio-temporal processes (PL-STP) \cite{tan2023temporal}\cite{tan2024openstl} with dynamic non-linearities and complex spatial domains has been increasingly critical and challenging in many scientific and engineering fields \cite{xu2020multi}\cite{degrave2022magnetic}. The essence of PL-STP can be mathematically defined as operator, which is formulated as constructing the mappings between infinite dimensional functions \cite{wang2023scientific}\cite{azzizadenesheli2023neural}, including temporal, spatial, and spatio-temporal functions. {Prominent example of constructing operators for PL-STP is solving solution functions (spatio-temporal functions) of parametric Partial Differential Equations (PDEs) \cite{li2020neural}\cite{yuan2023machine}\cite{huang2024lordnet} based on different initial conditions (spatial functions). In addition, there are many practical applications, such as predicting curing process-induced deformation field (spatial functions) for composite manufacturing based on temperature field (spatio-temporal functions) \cite{liu2023active} and predicting blood flow velocity field (spatio-temporal functions) for medical diagnosis and treatment based on blood pressure signals (temporal functions) \cite{maul2023transient}.} Typically, numerical simulations are the dominant approaches for PL-STP. However, for the complicated spatio-temporal process, especially for those with complex transient characteristics and spatial domains, the expensive computational cost of numerical methods remains prohibitive for real-time prediction and many-query analyses \cite{gao2021phygeonet}. 

With the advance of Artificial Intelligence (AI), deep learning has emerged as an alternative paradigm for PL-STP, which learns operators directly from observational data without knowledge of underlying PDE \cite{wang2023scientific}. In 2019, Lu et al. \cite{lu2021learning} extended the universal approximation theorem of Chen \cite{chen1995universal} and proposed the first deep operator network DeepONet, which employs two sub-networks to encode the locations of the output functions (trunk net) and sensors of the input functions (branch net), respectively. {Moreover, several tentative methods with similar sub-network structures have been proposed to learn the latent representations of dynamical systems \cite{regazzoni2023latent}\cite{pan2023neural}\cite{hua2023basis}. However, when it comes to real-world 3D time-dependent output functions, these methods would struggle with massive spatio-temporal locations due to the point-wise training, preventing them from acquiring accurate predictions \cite{lu2022multifidelity}.} To alleviate this constraint, Lu et al. \cite{lu2022comprehensive} developed a new extension of DeepONet, called POD-DeepONet, which replaced the trunk net with the bases (modes) calculated on training data by proper orthogonal decomposition (POD) \cite{lumley1967structure}. {Recently, such dimensionality reduction paradigms, including linear methods, such as POD \cite{fries2022lasdi} and principal component analysis (PCA) \cite{bhattacharya2021model}, and nonlinear methods, such as multilayer perceptrons (MLPs) \cite{seidman2022nomad}, kernel PCA \cite{eivazi2024nonlinear} and convolutional autoencoders (CAE) \cite{li2023latent}, have been advocated in operator learning due to the efficient description of systems evolution in low-dimensional latent spaces. Nevertheless, these approaches are either problem-specific or generic but moderate. }For instance, PCA-Net presented by Bhattacharya et al. \cite{bhattacharya2021model} conducts PCA-based \cite{wold1987principal} dimensionality reduction on input and output spaces and employs a fully-connected neural network (FC-NN) to construct mappings between two finite-dimensional latent spaces. Despite a generic framework of operator learning, the favourable performance of PCA-Net comes only with excellent reconstruction accuracy of truncated bases for the outputs, which is hard to achieve with a relatively small truncation number, especially when the outputs are complex spatio-temporal functions. In practice, increasing the number of truncated bases (i.e. increasing the complexity of neural networks) may only result in a marginal gain in accuracy or even a decrease in accuracy owing to increased mapping dimension \cite{jha2024residual}. 

Additionally, there is another promising operator learning architecture, Neural Operators (NOs). In contrast to conventional neural networks, in which the network parameterisation heavily relies on the discretisation of the inputs and outputs, NOs aim to approximate operators through a discretisation-invariant network structure \cite{kovachki2021neural}. As a result, NOs trained on a specific discretisation-resolution can be generalised to other discretisation-resolution without re-training. Li et al. \cite{li2020neural} first proposed Fourier Neural Operator (FNO) for learning operators between two same domains, which is inspired by spectral methods for solving differential equations \cite{trefethen1996finite}. As the name implies, FNO utilizes Fast Fourier Transformation (FFT) to encode the original function into Fourier domain and parameterises it there. Over the past few years, FNO and its variants, such as Wavelet Neural Operator (WNO) \cite{tripura2022wavelet} and U-shaped Neural Operator (UNO) \cite{rahman2022u}, have shown impressive results in operator learning of temporal and regular spatial domains \cite{chen2023physics,gopakumar2023fourier}. However, most real-world scenarios have complex spatial domains and use unstructured mesh for discrete representation, where the Fourier transformation in FNO, Wavelet transform in WNO, and image convolution in UNO designed for the uniform grid cannot be directly applied \cite{lu2022comprehensive,li2022fourier}. {Accordingly, Li et al. \cite{li2022fourier} further investigated the Geo-FNO model, which first learns to deform the irregular physical domain into a latent space with a uniform grid and then applies the FNO in the latent space. However, the application of Geo-FNO to general topologies would encounter limitations because the required diffeomorphism is challenging for typical engineering problems.} \textcolor{black}{To tackle the gap, our team reported a new concept, Neural Operator on Riemannian Manifolds (NORM) \cite{Chen2023LearningNO}, and generalise the NOs from Euclidean spaces to Riemannian manifolds by introducing the Laplacian eigenfunctions \cite{aflalo2015optimality,patane2018laplacian}}. NORM converts the function-to-function mapping into a finite-dimensional mapping in the subspace of the Laplacian eigenfunctions of the geometry while holding universal approximation property and preserving the discretisation-independent model structure. 

\textcolor{black}{Although the aforementioned NOs like NORM and FNO have achieved promising progress in operator learning, their similar Iterative Kernel Integration (IKI) based encoder-approximator-decoder blocks \cite{kovachki2021neural} determine that they require input space $D_{in}$ and output space $D_{out}$ to be the same domain \cite{lu2022comprehensive}, e.g. $D_{in}=D_{out}$. For example, FNO \cite{li2020fourier} utilizes 2D FFT and its inverse IFFT as encoder and decoder, restricting the $D_{in}$ and $D_{out}$ to be the same 2D spatial domain. However, many scenarios in PL-STP have unequal input and output domains, such as predicting the deformation field based on the temperature field for composite manufacturing, where the inputs are spatio-temporal functions (defined on both spatial and temporal domains) and the outputs are spatial functions (only defined on spatial domain). In this paper, we refer to these scenarios as unequal-domain mappings in PL-STP and divide them into two categories: }
\textcolor{black}{
\begin{itemize}
    \item[$\bullet$]	Increase-domain mapping: The output space $D_{out}$ is a product space of the input space $D_{in}$ and another space $D_e$ , i.e., $D_{out}=D_{in}\times D_e$.
    \item[$\bullet$]	Decrease-domain mapping: The input space$ D_{in}$ is a product space of the output space $D_{out}$ and another space $D_e$ , i.e., $D_{in}=D_{out}\times D_e$.
\end{itemize}} 
\textcolor{black}{Once confronted with unequal-domain mappings, IKI-based NOs have to perform expansion (for increase-domain mapping) or shrinkage (for decrease-domain mapping) at the input or middle layer to match the input and output domains \cite{lu2022comprehensive}, which brings difficulties in ensuring the predictive performance and stability of operator learning. Besides, it has been mentioned above that DeepONet and its variants will be limited by point-wise training, and dimensionality reduction methods will be generic but moderate, especially when dealing with increase-domain mapping. Consequently, it is necessary to fill the requirements of learning operator for unequal-domain mappings in PL-STP.}

\begin{figure}[t] 
\centering
\includegraphics[width=0.9\textwidth]{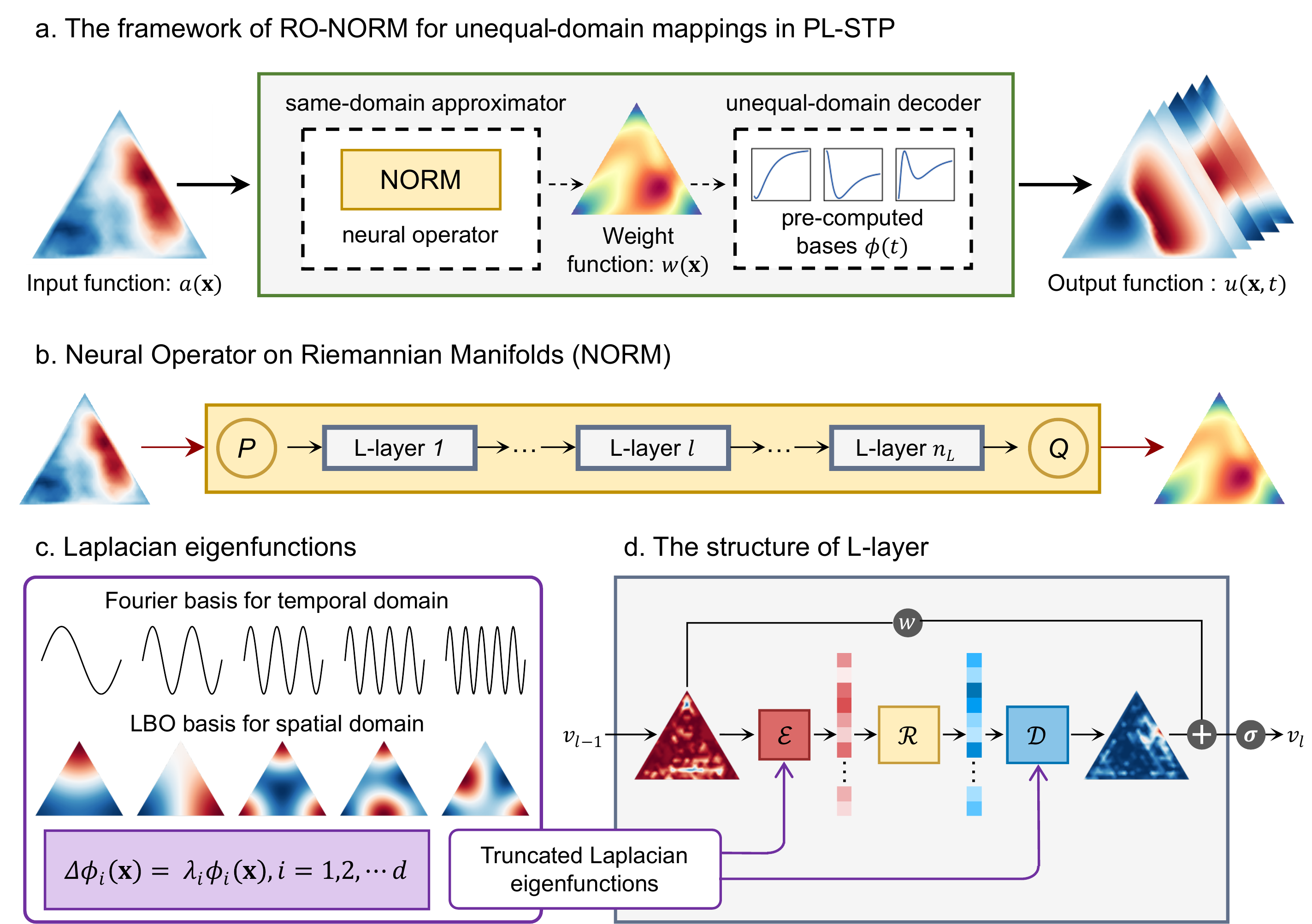}
\caption{\textcolor{black}{The illustration of the proposed RO-NORM. \textbf{a}, The framework of RO-NORM for unequal-domain mappings in PL-STP. \textbf{b}, The framework of NORM. \textbf{c}, The Laplacian eigenfunctions considered in L-layer, including Fourier basis for temporal domain and LBO basis for spatial domain. \textbf{d}, The structure of L-layer.}}
\label{Fig1}
\end{figure}

\textcolor{black}{This paper proposes a general reduced-order neural operator named Reduced-Order Neural Operator on Riemannian Manifolds (RO-NORM), consisting of two parts: unequal-domain encoder/decoder and same-domain approximator. }Motivated by the separation of variables in classical modal decomposition to analyse spatio-temporal patterns \cite{taira2017modal}, unequal-domain encoder/decoder transforms the spatio-temporal function $u(\mathbf{x}, t)$ into a sum of products between spatial $\phi(\mathbf{x})$ (or temporal $\phi(t)$) bases and corresponding temporally $w(t)$ (or spatially $w(\mathbf{x})$) distributed weight functions. Because the bases can be pre-computed, the original unequal-domain mapping is reformulated as the same-domain mapping between the spatial (or temporal) weight function and the spatial (or temporal) function of input or output, as shown in Fig.  \ref{Fig1}. Then the same-domain approximator NORM is applied to model the transformed mapping. Compared to the vector mapping simplified by classical dimensionality reduction methods, the mapping transformed by RO-NORM maintains the infinite property of the operator, which is no longer limited to modelling with FC-NN. Compared to NORM, RO-NORM transforms the unequal-domain mappings into same-domain mappings so that NORM can be applied directly without performing expansion and shrinkage to match domains. Meanwhile, benefiting from the reduced-order idea, RO-NORM has an advantage over existing IKI-based NOs in terms of training efficiency. Various comparative experiments, including learning solution operators of parametric PDEs, composite curing temperature and deformation field prediction, and aortic blood flow velocity prediction, have demonstrated the superiority of RO-NORM in prediction accuracy and training efficiency for spatio-temporal predictive learning.

\textcolor{black}{
\section{Reduced-Order Neural Operator on Riemannian Manifolds}
\subsection{Problem formulation}
}

\textcolor{black}{Consider the input function $a$ defined on arbitrary spatial domain $\mathcal{M}$ with boundary $\partial\mathcal{M}$ and temporal domain $\mathcal{T}$, and takes values in $\mathbb{R}^{d_a}$. Suppose the Hilbert space $\mathcal{A}$ defined with norm and inner product as the function space for the input function so that $a\in\mathcal{A}\left(\mathcal{M},\mathcal{T};\mathbb{R}^{d_a}\right)$. For most practical problems, the input function $a$ is the scalar function, so $d_a=1$. Three categories of functions can be sampled from $\mathcal{A}\left(\mathcal{M},\mathcal{T};\mathbb{R}^{d_a}\right)$, namely the standard spatio-temporal functions $a\left(\mathbf{x},t\right)$,}
\textcolor{black}{
\begin{equation}
    a\left(\mathbf{x},t\right):\ \ \left(\mathcal{M},\mathcal{T}\right)\rightarrow\mathbb{R}^{d_a}\ \ \ \ \mathbf{x}\in\mathcal{M},\ t\in\mathcal{T}
\end{equation}
the spatial functions $a\left( \mathbf{x} \right)$ at a specific moment (e.g. initial condition $a\left(x,t=0\right)$)
\begin{equation}
    a\left( \mathbf{x} \right):\ \ \mathcal{M}\rightarrow\mathbb{R}^{d_a}\ \ \ \ \mathbf{x}\in\mathcal{M}
\end{equation}
and temporal functions $a\left( t \right)$ with fixed location (e.g. boundary condition $a\left(x=\partial\mathcal{M},t\right)$).
\begin{equation}
    a\left( t \right):\ \ \mathcal{T}\rightarrow\mathbb{R}^{d_a}\ \ \ \ t\in\mathcal{T}
\end{equation}}\textcolor{black}{
Similarly, define $u\in\mathcal{U}\left(\mathcal{M},\mathcal{T};\mathbb{R}^{d_u}\right)$ as the output function and the $\mathcal{G}:\ \mathcal{A}\left(\mathcal{M},\mathcal{T};\mathbb{R}^{d_a}\right)\rightarrow\mathcal{U}\left(\mathcal{M},\mathcal{T};\mathbb{R}^{d_u}\right) $ as the target operator mapping. This paper focuses on unequal-domain mappings in PL-STP, which pose challenges to existing operator learning methods. Specifically, there are a total of two kinds of increase-domain mappings: 
\begin{equation}
\begin{aligned}
    &\mathcal{G}:~a\left( \mathbf{x} \right)\rightarrow u\left( {\mathbf{x},t} \right),~~~~a \in A\left( {\mathcal{M};\mathbb{R}^{d_{a}}} \right),~~~u \in U\left( {\mathcal{M},\mathcal{T};\mathbb{R}^{d_{u}}} \right) \\
    &\mathcal{G}:~a\left(t\right)\rightarrow u\left( {\mathbf{x},t} \right),~~~~a \in A\left( {\mathcal{T};\mathbb{R}^{d_{a}}} \right),~~~u \in U\left( {\mathcal{M},\mathcal{T};\mathbb{R}^{d_{u}}} \right) \\
\end{aligned}
\end{equation}}
\textcolor{black}{and two kinds of decrease-domain mappings:\begin{equation}
\begin{aligned}
    &\mathcal{G}:~a\left(\mathbf{x},t\right)\rightarrow u\left(\mathbf{x}\right),~~~~a \in A\left( {\mathcal{M},\mathcal{T};\mathbb{R}^{d_{a}}} \right),~~~u \in U\left( {\mathcal{M};\mathbb{R}^{d_{u}}} \right) \\
    &\mathcal{G}:~a\left( {\mathbf{x},t} \right)\rightarrow u\left(t\right),~~~~a \in A\left( {\mathcal{M},\mathcal{T};\mathbb{R}^{d_{a}}} \right),~~~u \in U\left( {\mathcal{T};\mathbb{R}^{d_{u}}} \right)
\end{aligned}
\end{equation}}

\textcolor{black}{The goal of the neural operator is to approximate the $\mathcal{G}$ by learning a parametric operator $\mathcal{G}_\theta$ with a network parameterised by $\theta\in\mathbb{R}^p$, which equals to solving the minimisation problem:
\begin{equation}
    \underset{\theta \in \mathbb{R}^{p}}{min}\,\mathbb{E}{\parallel {\mathcal{G} - \mathcal{G}_{\theta}} \parallel}_{\mathcal{U}}
\end{equation}}

\textcolor{black}{Given the training data $\left\{a_i, u_i\right\}_{i=1}^N$, we can reformulate the equation as the empirical-risk minimisation problems:
\begin{equation}
\underset{\theta \in \mathbb{R}^{p}}{min}\,\frac{1}{N}{\sum\limits_{i = 1}^{N}\,}{\parallel {u_{i} - \mathcal{G}_{\theta}\left( a_{i} \right)} \parallel}_{L^{2}}
\end{equation}}

\textcolor{black}{Considering that the temporal and spatial dimensions play the same role in each category, without loss of generality, we take $\mathcal{G}:\ a\left(\mathbf{x}\right)\rightarrow u\left(\mathbf{x},t\right)$ and $\mathcal{G}:\ a\left(\mathbf{x},t\right)\rightarrow u\left(\mathbf{x}\right)$ as the examples of increase-domain mapping and decrease-domain mapping respectively, in the following.}

\textcolor{black}{\subsection{Unequal-domain encoder and decoder}}

\textcolor{black}{As mentioned in the Introduction, although the IKI-based NOs work well in same-domain mapping, namely $a\left(t\right)\rightarrow u\left(t\right)$ and $a\left(\mathbf{x}\right)\rightarrow u\left(\mathbf{x}\right)$, they are incapable of learning unequal-domain mapping directly and have to perform expansion or shrinkage to match the input and output domains. On the other hand, dimensionality reduction methods such as PCA-Net have more general architecture by transforming mappings between infinite-dimensional functions into mappings between finite-dimensional vectors and using FC-NN for modelling. However, their dimensionality reduction strategy on the entire function space may cause moderate results when facing complex spatio-temporal functions \cite{holmes2012turbulence}. Instead, RO-NORM designs an unequal-domain encoder/decoder with separation of variables for dimensionality reduction. We employ a set of bases to perform dimensionality reduction against the extra space $D_e$ in the input or output, e.g. using a set of temporal bases $\phi\left(t\right)$ to reduce the time dimension of the $u\left(\mathbf{x},t\right)$ and $a\left(\mathbf{x},t\right)$. Then, the spatio-temporal function $u\left(\mathbf{x},t\right)$ (or $a\left(\mathbf{x},t\right)$) is formulated into a sum of products between temporal bases $\phi\left(t\right)$ and corresponding spatially distributed weight functions $w\left(\mathbf{x}\right)$. Since the bases can be pre-computed, the original unequal-domain mapping can be transformed into the same-domain mapping between the spatial weight function $w\left(\mathbf{x}\right)$ and the spatial function of input $a\left(\mathbf{x}\right)$ or output $u\left(\mathbf{x}\right)$. }

\textcolor{black}{This section mathematically describes how to construct the unequal-domain encoder/decoder by dimensionality reduction with the separation of variables and how to choose the bases.}

\textcolor{black}
{
\subsubsection{Unequal-domain encoder and decoder with dimensionality reduction}
\label{unequal-domain-encoder-and-decoder-with-dimensionality-reduction}
}

\textcolor{black}{To construct the unequal-domain encoder/decoder for the task
\(a\left( \mathbf{x} \right) \rightarrow u\left( \mathbf{x},t \right)\)
(or
\(\mathcal{G:\ }a\left( \mathbf{x},t \right) \rightarrow u\left( \mathbf{x} \right)\)),
temporal bases are needed to reduce the dimensionality of the extra
temporal domain in \(u\left( \mathbf{x},t \right)\) (or
\(a\left( \mathbf{x},t \right)\)). Denote the
\(u_{i}\left( \mathbf{x},t \right)\mathcal{\in U}\left( \mathcal{M,T;}\mathbb{R}^{d_{u}} \right)\)
the \(i\)th sample of training data \({\{ a_{i},\ u_{i}\}}_{i = 1}^{N}\)
and
\(u_{i,j}(t)\mathcal{\in U}\left( \mathcal{T;}\mathbb{R}^{d_{u}} \right)\)
for the \(j\)th spatial point evolution of
\(u_{i}\left( \mathbf{x},t \right)\), which is a temporal function.
Then, the spatio-temporal function \(u_{i}\left( \mathbf{x},t \right)\)
can be represented as the temporal function of each spatial location:}

\textcolor{black}{
\begin{equation}
    u_{i}\left( {\mathbf{x},t} \right) = \left( {u_{i,1}\left(t\right),u_{i,2}(t),\cdots,u_{{i,n}_{x}}\left(t\right)} \right),~~u_{i,j}\left(t\right) \in \mathcal{U}\left( {\mathcal{T};\mathbb{R}^{d_{u}}} \right)
    \label{8}
\end{equation}}\textcolor{black}{
where \(n_{x}\) means the total number of spatial locations. Naturally,suppose
\(\phi_{1,\mathcal{T}},\ \phi_{2,\mathcal{T}},\ \cdots,\phi_{d,\mathcal{T}} \in \mathcal{U}\left( \mathcal{T;}\mathbb{R}^{d_{u}} \right)\)
are a set of temporal orthogonal bases. Therefore, we can project the
temporal \(u_{i,j}\left(t\right)\) in \(u_{i}\left( \mathbf{x},t \right)\) onto
\(d\)-dimensional sub-space through the bases. For any \(d \geq 1,\) we
can define a \(d\)-dimensional projection subspace \(\mathcal{U}^{'}\)
of \(\mathcal{U}\left( \mathcal{T;}\mathbb{R}^{d_{u}} \right)\),}

\textcolor{black}{
\begin{equation}
    \mathcal{U}^\prime=span\left\{\phi_{1,\mathcal{T}},\phi_{2,\mathcal{T}},\cdots,\phi_{d,\mathcal{T}}\right\}\subset\mathcal{U}\left(\mathcal{T};\mathbb{R}^{d_u}\right)
\end{equation}}

\textcolor{black}{
Define the sub-encoder \(\Pi_E:\ \mathcal{U}\rightarrow\mathbb{R}^d\) as the mapping from \(\mathcal{U}\left(\mathcal{T};\mathbb{R}^{d_u}\right)\) to the coefficients of the orthogonal projection onto \(\mathcal{U}^\prime\), that is
\begin{equation}
\Pi_E\left(u_{i, j}\left(t\right)\right):=\left(\left\langle u_{i, j}\left(t\right), \phi_{1, T}\right\rangle, \cdots,\left\langle u_{i, j}\left(t\right), \phi_{d, T}\right\rangle\right)
\end{equation}
}

\textcolor{black}{
With the sub-encoder \(\Pi_E\) of temporal sequence \( u_{i,j}\left(t\right)\), we can define the extended encoder \(\mathcal{F}_E \) on \(u_i\left(\mathbf{x},t\right)\)
\begin{equation}
    \mathcal{F}_{E}\left( {u_{i}\left( {\mathbf{x},t} \right)} \right) = \left( {\Pi_{E}\left( {u_{i,1}\left(t\right)} \right),\Pi_{E}\left( {u_{i,2}\left(t\right)} \right),\cdots,\Pi_{E}\left( {u_{{i,n}_{x}}\left(t\right)} \right)} \right),~~~~\Pi_{E}\left( {u_{i,j}\left(t\right)} \right) \in \mathbb{R}^{d}
\end{equation}
}\textcolor{black}{
The proposed \(\mathcal{F}_{E}\) can encode the spatio-temporal function
\(u_{i}\left( \mathbf{x},t \right)\) as the spatial weight function that
takes values in \(\mathbb{R}^{d}\), namely
\(\mathcal{F}_{E}:\mathcal{U}\left( \mathcal{M,T;}\mathbb{R}^{d_{u}} \right)\mathcal{\rightarrow W(M;}\mathbb{R}^{d})\).
}

\textcolor{black}{
Additionally, we define the inverse sub-decoder
\(\Pi_{d}:\ \mathbb{R}^{d}\mathcal{\rightarrow U}\) that reconstructs
the elements in
\(\mathcal{U}\left( \mathcal{T;}\mathbb{R}^{d_{u}} \right)\) by a sum of
products between its weight coefficients and pre-calculated bases
\(\phi_{1,\mathcal{T}},\ \phi_{2,\mathcal{T}},\ \cdots,\phi_{d,\mathcal{T}}\),
namely
}

\textcolor{black}{
\begin{equation}
    \ \Pi_{d}(w) := \sum_{k = i}^{d}w_{k}\phi_{k,\mathcal{T}}\ \ \ \ \ \forall w \in \mathbb{R}^{d}
\end{equation}
}

\textcolor{black}{
With the sub-decoder \(\Pi_{d}\), we can define the unequal-domain
decoder \(\mathcal{F}_{D}\) on spatial weight function
\(w_{i}\left( \mathbf{x} \right)\):
}

\textcolor{black}{
\begin{equation}
    \mathcal{F}_{D}\left( w_{i}\left( \mathbf{x} \right) \right) = \left\lbrack \Pi_{d}\left( w_{i,1} \right),\Pi_{d}\left( w_{i,2} \right),\cdots,\Pi_{d}\left( w_{{i,n}_{x}} \right) \right\rbrack,\ \ \ \ w_{i,j} \in \mathbb{R}^{d},\ \Pi_{d}\left( w_{i,1} \right) \in \mathcal{U}\left( \mathcal{T};\mathbb{R}^{d_{u}} \right)
\end{equation}
}\textcolor{black}{
The proposed \(\mathcal{F}_{D}\) can decode the spatial weight function
\(w_{i}\left( \mathbf{x} \right)\) that takes values in
\(\mathbb{R}^{d}\) as the spatio-temporal function, namely
\(\mathcal{F}_{D}:\mathcal{W(M;}\mathbb{R}^{d}\mathcal{) \rightarrow U}\left( \mathcal{M,T;}\mathbb{R}^{d_{u}} \right)\).
}

\textcolor{black}{With the unequal-domain encoder \(\mathcal{F}_{E}\), we can reformulate
the decrease-domain mapping
\(\mathcal{G:}a\left( \mathbf{x},t \right) \rightarrow u\left( \mathbf{x} \right)\)
as:}

\textcolor{black}{
\begin{equation}
    \mathcal{G}:\ a\left( \mathbf{x} \right)\overset{\mathcal{G}^{'}}{\rightarrow}w\left( \mathbf{x} \right)\overset{\mathcal{F}_{D}}{\rightarrow}u\left( \mathbf{x},t \right)
\end{equation}
}

\textcolor{black}{With the unequal-domain encoder \(\mathcal{F}_{E}\), we can reformulate
the decrease-domain mapping
\(\mathcal{G:}a\left( \mathbf{x},t \right) \rightarrow u\left( \mathbf{x} \right)\)
as:}

\textcolor{black}{
\begin{equation}
   \mathcal{G}:\ a\left( \mathbf{x},t \right)\overset{\mathcal{F}_{E}}{\rightarrow}w\left( \mathbf{x} \right)\overset{\mathcal{G}^{'}}{\rightarrow}u\left( \mathbf{x} \right)
\end{equation}
}

\textcolor{black}{
Since the \(\mathcal{F}_{E}\) and \(\mathcal{F}_{D}\) are
pre-calculated, the original unequal-domain mapping \(\mathcal{G}\) can
be transformed into the same-domain mapping \(\mathcal{G}^{'}\). In what
follows, we attempt to analyse the choice of basis.
}

\textcolor{black}{
\subsubsection{Basis for dimensional reduction}
\label{basis-for-dimensional-reduction}
}

\textcolor{black}{
The core of the unequal-domain encoder/decoder is to leverage a set of
bases to deconstruct/reconstruct the extra space \(D_{e}\) in the input
or output. This section aims to introduce the choice of basis. The
optional bases can be divided into two categories: data-dependent bases,
which are computed from the data, and data-independent bases, which are
intrinsic to the domain.}

\textcolor{black}{\textbf{(1) Data-dependent bases: Proper orthogonal decomposition bases}}

\textcolor{black}{
One of the widely used data-dependent bases is extracted by proper
orthogonal decomposition, which was first proposed for fluid dynamics
analysis \cite{lumley1967structure}. Given equation \ref{8}, we can consider the empirical,
non-centred covariance operator.
}

\textcolor{black}{
\begin{equation}
C := \frac{1}{N \times n_{x}}{\sum\limits_{i = 1}^{N}\,}{\sum\limits_{j = 1}^{n_{x}}\,}_{i,j}u_{i,j}\left(t\right) \otimes u_{i,j}\left(t\right)
\end{equation}
}\textcolor{black}{
where \(\otimes\) denotes the outer product. Through eigenvalue
decomposition, a sequence of eigenvalues
\(\lambda_{1,\mathcal{T}} \geq \lambda_{2,\mathcal{T}} \geq \cdots \geq 0\)
and corresponding eigenvectors
\(\phi_{1,\mathcal{T}},\ \phi_{2,\mathcal{T}},\ \cdots \in \mathcal{U}\left( \mathcal{T;}\mathbb{R}^{d_{u}} \right)\),
i.e. POD bases can be obtained. In Section 3, we will apply POD bases to
reduce the dimension of the extra spatial or temporal domain.
}

\textcolor{black}{\textbf{(2) Data-independent bases: Laplacian eigenfunctionsl decomposition bases}}

\textcolor{black}{The Laplacian eigenfunctions \(\phi\left( \mathbf{x} \right)\) are
obtained by solving the eigenvalue problem for the Laplacian
\(\mathrm{\Delta}\) on a bounded domain \(\Omega\) with boundary
\(\partial\Omega\), where the following condition is satisfied
}

\textcolor{black}{
\begin{equation}
\Delta\phi\left( \mathbf{x} \right) = \ \lambda\phi\left( \mathbf{x} \right),\ \ \mathbf{x}\mathbf{\in} \Omega
\end{equation}
}\textcolor{black}{
in addition to the Dirichlet boundary condition
\(\phi\left( \mathbf{x} \right) = f_{0}\left(\mathbf{x}\right),\ \ \mathbf{x \in}\partial\Omega\),
Neumann boundary conditions
\(\nabla\phi\left( \mathbf{x} \right) \bullet \widehat{n} = g_{0}\left(\mathbf{x}\right),\ \ \mathbf{x \in}\partial\Omega\)
can also be considered \cite{aflalo2015optimality}, where \(\widehat{n}\) indicates the
normal direction of the boundary. The \(f_{0}(\mathbf{x})\) and
\(g_{0}(\mathbf{x})\) are often zero \cite{reuter2006laplace}. The
\(\lambda_{1} \leq \lambda_{2} \leq \cdots\) and
\(\phi_{1}\left( \mathbf{x} \right),\phi_{2}\left( \mathbf{x} \right),\cdots\)
are Laplacian eigenvalues and the corresponding eigenfunctions.
Moreover, with the definition of the divergence operator
\(\nabla \bullet\) and gradient operator \(\nabla f\) on manifolds with
Riemannian metric \(g\), the Laplacian can be naturally extended to the
Riemannian manifolds, called Laplace--Beltrami operator (LBO) \cite{reuter2006laplace}.
Theoretical research has demonstrated that LBO eigenfunctions are a set
of optimal orthonormal basis that can approximate functions defined on
Riemannian manifolds with any accuracy \cite{aflalo2015optimality}.
}

\textcolor{black}{
As for the temporal domain, Fourier bases are the typical intrinsic
bases, which are closely related to the Fourier transform and are
usually considered effective representations of functions in engineering
applications, such as various signals \cite{rust2013convergence}. Actually, some studies
have pointed out that the Laplace eigenfunctions are the generalization
of Fourier bases \cite{wang2019intrinsic}\cite{Tererce2016fourier}. In summary, we can use the intrinsic
LBO and Fourier bases to compress the extra spatial and temporal
domains, respectively, which will be discussed in Section 4.4.
}

\textcolor{black}{
\subsection{Same-domain approximator with
NORM}
\label{same-domain-approximator-with-norm}
}

\textcolor{black}{In this section, we apply NORM as the same-domain approximator to model
the transformed mapping in Section 2.2.1:
\(\mathcal{G}^{'}:a\left( \mathbf{x} \right) \rightarrow w\left( \mathbf{x} \right)\)
and
\(\mathcal{G}^{'}:w\left( \mathbf{x} \right) \rightarrow u\left( \mathbf{x} \right)\).}
The core architecture of the NORM lies in the Laplacian kernel integral
operator, denoted as
\(\mathcal{N}_{\theta}\mathcal{= D \circ}\mathcal{R}_{\theta}\mathcal{\circ E}\),
which consists of three-step procedures: the encoder \(\mathcal{E}\),
the approximator \(\mathcal{R}_{\theta}\) and the decoder
\(\mathcal{D}\), as shown in Fig. \ref{Fig1} (d).

Firstly, the encoder \(\mathcal{E}\) projects the input function
\(a(x)\) to Laplacian spectrum, which can be defined as the spectral
decomposition on the LBO eigenfunctions \(\phi_{i, \mathcal{M}}\) of
the input manifold \(\mathcal{M}\):

\begin{equation}
    \mathcal{E}:\mathcal{A}\left( \mathcal{M};\mathbb{R}^{d_{a}} \right) \rightarrow \mathbb{R}^{d_{\mathcal{M}}}\mathcal{,\ \ \ \ \ E}(a) := \left( \left\langle a,\phi_{1, \mathcal{M}} \right\rangle,\ldots,\left\langle a,\phi_{d_{\mathcal{M}}, \mathcal{M}} \right\rangle \right),\ \ \ \forall a \in \mathcal{A}
\end{equation}

After that, a finite-dimensional parameterisation Euclidean mapping
\(\mathcal{R}_{\theta}\) is applied to spectral coefficients in
\(\mathbb{R}^{d_{\mathcal{M}}}\). And the number of parameters of
\(\mathcal{R}_{\theta}\) only relies on the size of the truncated
Laplacian eigenfunctions \(d_{\mathcal{M}}\), which reflects the
discretisation-invariant properties.

\begin{equation}
    \mathcal{R}_{\theta}:\mathbb{R}^{d_{\mathcal{M}}} \rightarrow \mathbb{R}^{d_{\mathcal{M}}}
\end{equation}

Lastly, the parameterised coefficients can be mapped back to the
\(\mathcal{W}\) by the decoder \(\mathcal{D}\), which equals the
spectral reconstruction on the LBO eigenfunctions:

\begin{equation}
    \mathcal{D}:\mathbb{R}^{d_{\mathcal{M}}} \rightarrow \mathcal{W}\left( \mathcal{M};\mathbb{R}^{d} \right),\ \ \ \ \mathcal{D}(\beta) = \sum_{i = 1}^{d_{\mathcal{M}}}\mspace{2mu}\beta_{i}\phi_{i,\ \mathcal{M}}\ \ ,\ \ \ \ \ \forall\beta \in \mathbb{R}^{d_{\mathcal{M}}}
\end{equation}

To overcome the inefficiency of modelling complex nonlinear operators
with one Laplacian kernel integral operator, NORM adopts a deep
iterative framework comprised of the lifting layer \(\mathcal{P}\),
\(n_{L}\) kernel integral layers and the projecting layer
\(\mathcal{Q}\), as shown in Fig. \ref{Fig1} (b).

Similar to the convolution channel expansion in CNN, lifting layer
\(\mathcal{P:}\mathcal{A}\left( \mathcal{M;}\mathbb{R}^{d_{a}} \right)\mathcal{\rightarrow H}\left( \mathcal{M;}\mathbb{R}^{d_{w}} \right)\)
lifts the channels of the input function \(a\) to improve the
representation capability, where
\(\mathcal{H}\left( \mathcal{M;}\mathbb{R}^{d_{w}} \right)\) denotes
Hilbert space \(\mathcal{H}\) on Riemannian manifolds \(\mathcal{M}\)
with functions taking values in \(\mathbb{R}^{d_{w}}\). Multiple kernel
integral layers, defined as Laplace layer, or L-layer, update the input
function iteratively as
\(v_{1} \mapsto v_{2} \mapsto \ldots \mapsto v_{L}\). And the iteration
in L-layer \(l\) is given as follows:

\begin{equation}
v_{l} = \mathcal{L}_{l}\left( v_{l - 1} \right)\left( \mathbf{x} \right):= \sigma\left( {W_{l}v_{l - 1}\left( \mathbf{x} \right) + b_{l}\left( \mathbf{x} \right) + \mathcal{N}_{\theta}\left( v_{l - 1} \right)\left( \mathbf{x} \right)} \right),~~\forall x \in \mathcal{M}
\end{equation}
where the linear transformations
\(W_{l} \in \mathbb{R}^{d_{w} \times d_{w}}\) and the bias
\(b_{l} \in \mathbb{R}^{d_{w}}\) are defined as point-wise mapping.
\(\sigma\) is the common nonlinear activation function used in neural
networks. \(\mathcal{N}_{\theta}\) is the Laplacian kernel integral
operator. Finally, the projecting layer
\(\mathcal{Q:}\mathcal{H}\left( \mathcal{M;}\mathbb{R}^{d_{w}} \right)\mathcal{\rightarrow W(M;}\mathbb{R}^{d})\)
reduce the channels to the dimension of the output function.

\(\mathcal{P}\) and \(\mathcal{Q}\) are both learnable shallow networks,
which can implement the same lift or reduce operation for the channel of
all functions on \(\mathcal{M}\) regardless of
discretisation-resolution. For L-layer, in addition to the Laplacian
kernel integral operator \(\mathcal{N}\), \(W_{l}\) and \(b_{l}\) are
also discretisation-invariant since the parameterisation of them only
depends on the expanded channel \(d_{w}\) after \(\mathcal{P}\).
Consequently, NORM maintains the discretisation-invariant property of
NOs.

\textcolor{black}{
\subsection{Implementation of RO-NORM}
\label{implementation-of-ro-norm}
}

\textcolor{black}{
When implementing RO-NORM for the unequal-domain mapping in PL-STP, the
first step is obtaining a set of orthonormal bases to reduce the
dimension of the extra domain in the input or output. Optional bases
include data-dependent bases and data-independent bases, and the
difference between them will be discussed in Section 4.4. Then, the
unequal-domain encoder \(\mathcal{F}_{E}\) and the unequal-domain
decoder \(\mathcal{F}_{D}\) can be constructed. For increase-domain
mapping, a same-domain approximator NORM \(\mathcal{G}_{\theta}\) is
used to input \(a\left( \mathbf{x} \right)\) and output the prediction
of spatial weight
function\(\ w^{'}\left( \mathbf{x} \right) = \mathcal{G}_{\theta}\left(a\left(\mathbf{x}\right)\right)\),
and decoder \(\mathcal{F}_{D}\) is utilized to reconstruct the final
prediction
$u^\prime\left(\mathbf{x},t\right)=\mathcal{F}_D\left(w^\prime\left(\mathbf{x}\right)\right)$.
Depending on whether the reconstruction is considered in the training
process, namely, the training label is \(w\left( \mathbf{x} \right)\) or
\(u\left( \mathbf{x},t \right)\), the training pattern can be divided
into offline and online reconstruction, which will be discussed in
Section 4.5. For decrease-domain mapping, the original input
\(a\left( \mathbf{x},t \right)\) are firstly reduced into
\(w\left( \mathbf{x} \right)\) using encoder \(\mathcal{F}_{E}\). Then,
a same-domain approximator NORM \(\mathcal{G}_{\theta}\) is employed to
input \(w\left( \mathbf{x} \right)\) and output the final prediction
$u^\prime\left(\mathbf{x}\right)=\mathcal{G}_\theta\left(w\left(\mathbf{x}\right)\right)$.
The procedures of RO-NORM for increase-domain mapping and
decrease-domain mapping are sketched in Algorithms 1 (online
reconstruction scheme) and 2.
}

\begin{table}[htb]
\begin{tabular}{p{\linewidth}} 
\toprule
\textcolor{black}{Algorithm 1: RO-NORM for increase-domain mapping $a\left(\mathbf{x}\right)\rightarrow u\left(\mathbf{x},t\right)$ }\\
\midrule
Input: $N$-samples of the pair $\{a\left(\mathbf{x}\right)\in\mathcal{A}\left(\mathcal{M};\mathbb{R}^{d_a}\right),u\left(\mathbf{x},t\right)\in\mathcal{U}\left(\mathcal{M},\mathcal{T};\mathbb{R}^{d_u}\right)\}$ \\
1. Implement the separation of variables to obtain a set of orthonormal bases  $\phi_{1,\mathcal{T}},\phi_{2,\mathcal{T}},\cdots,\phi_{d,\mathcal{T}}$ of $\mathcal{U}\left(\mathcal{T};\mathbb{R}^{d_u}\right)$ and construct the encoder $\mathcal{F}_E$ and the decoder $\mathcal{F}_D$. \\
2. Input $a\left(\mathbf{x}\right)$ to NORM $\mathcal{G}_\theta$ and obtain the prediction of spatial weight function $w^\prime\left(\mathbf{x}\right)=\mathcal{G}_\theta\left(a\left( {\mathbf{x}} \right)\right)$. \\
3. Leverage the decoder $\mathcal{F}_D$ and $w^\prime\left(\mathbf{x}\right)$ to reconstruct the final prediction $u^\prime\left(\mathbf{x},t\right)=\mathcal{F}_D\left(w^\prime\left(\mathbf{x}\right)\right)$. \\
4. Solve the empirical-risk minimisation problem $\min_{\theta\in\mathbb{R}^p}\frac{1}{N}\sum_{i=1}^{N}\left\|u_i\left(\mathbf{x},t\right)-u_i^\prime\left(\mathbf{x},t\right)\right\|_{L^2}$. \\
Output: The NORM $\mathcal{G}_\theta$ and the decoder $\mathcal{F}_D$. $u\left(\mathbf{x},t\right)=\mathcal{F}_D\left(\mathcal{G}_\theta\left(a\left( {\mathbf{x}} \right)\right)\right)$. \\
\bottomrule
\end{tabular}
\end{table}

\begin{table}[htb]
\begin{tabular}{p{\linewidth}}
\toprule
\textcolor{black}{Algorithm 2: RO-NORM for decrease-domain mapping $a\left(\mathbf{x},t\right)\rightarrow u\left(\mathbf{x}\right)$} \\
\midrule
Input: $N$-samples of the pair $\{a\left(\mathbf{x},t\right)\in\mathcal{A}\left(\mathcal{M},\mathcal{T};\mathbb{R}^{d_a}\right),u\left(\mathbf{x}\right)\in\mathcal{U}\left(\mathcal{M};\mathbb{R}^{d_u}\right)\}$ \\
1. Implement the separation of variables to obtain a set of orthonormal bases  $\phi_{1,\mathcal{T}},\phi_{2,\mathcal{T}},\cdots,\phi_{d,\mathcal{T}}$ of $\mathcal{U}\left(\mathcal{T};\mathbb{R}^{d_u}\right)$ and construct the encoder $\mathcal{F}_E$ and the decoder $\mathcal{F}_D$. \\
2. Reduce the dimension of $a\left(\mathbf{x},t\right)$ via the encoder $\mathcal{F}_E$ to obtain the spatial weight function $w\left(\mathbf{x}\right)=\mathcal{F}_E\left(a\left(\mathbf{x},t\right)\right)$. \\
3. Input $w\left(\mathbf{x}\right)$ to NORM $\mathcal{G}_\theta$ and obtain the final prediction $u^\prime\left(\mathbf{x}\right)=\mathcal{G}_\theta\left(w\left(\mathbf{x}\right)\right)$. \\
4. Solve the empirical-risk minimisation problem $\min_{\theta\in\mathbb{R}^p}\frac{1}{N}\sum_{i=1}^{N}\left\| u_i\left(\mathbf{x}\right)-u_i^\prime\left(\mathbf{x}\right)\right\|_{L^2}$. \\
Output: The NORM $\mathcal{G}_\theta$ and the encoder $\mathcal{F}_E$. $u\left(\mathbf{x}\right)=\mathcal{G}_\theta \left(\mathcal{F}_E\left(a\left( {\mathbf{x},t} \right)\right)\right)$. \\
\bottomrule
\end{tabular}
\end{table}

\section{Numerical results}
In this section, we illustrate the capacity of the proposed RO-NORM on six benchmark cases, which cover two parametric PDE systems (i.e., 2D Burgers’ equations and 2D Wave equations) and four engineering and biomedical applications (i.e., the heat source layout prediction, composite workpiece temperature and deformation prediction, and blood flow dynamics prediction). Four baseline algorithms are compared with our method, including DeepONet, POD-DeepONet, PCA-Net, and vanilla NORM. \textcolor{black}{In each case, we keep the number of truncated modes the same for POD-DeepONet, PCA-Net and RO-NORM and guarantee that the percentage of kinetic energy exceeds 99$\%$. The details on the selection of hyperparameters, such as network structure, learning rate, epoch, batch size, etc, are provided in Table \ref{tbl12}.} The loss function for training is defined as the \(L_2\) relative error:
\begin{equation}
E_{L_2} = \frac{1}{N}{\sum\limits_{p = 1}^{N}\,}\frac{{\parallel {u_{p} - \hat{u_{p}}} \parallel}_{2}}{{\parallel u_{p} \parallel}_{2}}
\end{equation}
where $u_p$ denotes the ground truth, the $\widehat{u_p}$ is the prediction and $\|\cdot\|_2$ represents the $L_2$ norm. In addition to $L_2$ relative error, we consider the other metric called Mean Maximum Error (MME) for testing, which refers to the mean value of all test samples in terms of the maximum full-field error. The DeepONet and POD-DeepONet are implemented based on DeepXDE deep learning library \cite{lu2021deepxde}, and the rest of methods are implemented in PyTorch framework. The code and datasets for this work will become available at \href{https://github.com/qingluM/RO-NORM}{https://github.com/qingluM/RO-NORM}.

\subsection{2D Burgers’ equations}
The Burgers’ equation is a fundamental PDE arising in various areas, such as fluid mechanics and gas dynamics \cite{jagtap2020adaptive}. Furthermore, we consider the 2D viscous Burgers’ equation defined in a triangular domain, given in tensor form:
\begin{equation}
\mathbf{u}_t+\mathbf{u} \cdot \nabla \mathbf{u}-\beta \Delta \mathbf{u}=0, ~\mathbf{x}=(x, y) \in[0,1]^2, ~t \in[0,5]
\end{equation}
with periodic boundary conditions, where $\mathbf{u}=\left\{u, v\right\}$ is the fluid velocities, $\beta$ is the viscosity of the fluid.

\begin{figure}[t]
\centering
\includegraphics[width=0.85\textwidth]{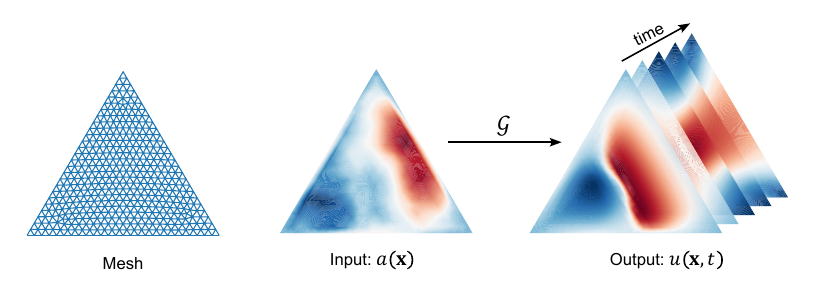}
\caption{Illustration of the 2D Burgers’ equations.}
\label{Fig2}
\end{figure}

\noindent \textbf{Problem setup.} Herein, we attempt to learn the operator mapping between the initial condition $a\left( \mathbf{x} \right) = u\left( {\mathbf{x},t=0} \right)$ and the spatio-temporal solution $u\left( {\mathbf{x},t} \right)$ for $t \in [0.05, 5]$, as shown in Fig. \ref{Fig2},
\begin{equation}
\left. \mathcal{G}:\mathcal{~}a\left( \mathbf{x} \right)\rightarrow u\left( {\mathbf{x},t} \right) \right.
\end{equation}

\begin{table}[b]
\caption{\textcolor{black}{Comparison of MME and $L_2$ relative error between RO-NORM and baseline methods on Burgers' equation ($a\left( \mathbf{x} \right)\rightarrow u\left( {\mathbf{x},t} \right)$). All the results are averaged over three repeated runs, including mean value and standard deviation. Results in bold are the best, and those with \underline{underlines} are the second best. '\textbf{\textit{Improvements}}' is the percentage of the best over the second best.}}
\label{tbl1}
\begin{tabular}{lccccc|c}
\toprule
Metrics & DeepONet & POD-DeepONet & PCA-Net & NORM & RO-NORM & \textbf{\textit{Improvements}} \\
\midrule
MME & \textcolor{black}{0.510 (0.001)} & \textcolor{black}{0.233 (0.001)} & 0.209 (0.001) & \underline{0.204 (0.015)} & \textbf{0.065 (0.001)} & \textbf{\textit{68.13\%}} \\
$E_{L_2}(\%)$ & \textcolor{black}{52.251 (0.084)} & \textcolor{black}{19.582 (0.010)} & 18.117 (0.076) & \underline{15.415 (1.529)} & \textbf{4.356 (0.066)} & \textbf{\textit{71.74\%}} \\
\bottomrule
\end{tabular}
\end{table}

\noindent \textbf{Data generation.} We follow the source code in \cite{li2020fourier} to generate the initial conditions from \textcolor{black}{a Gaussian random field. \(GRF\  = \ GaussianRF(dim = 2,\ s = 96,\ alpha = 3,\ tau = 3,\ sigma = None,\ boundary = periodic)\).} Then we choose $\beta=0.005$ and use the commercial simulation software COMSOL to calculate the ground truth solutions $u\left(\mathbf{x},t\right)$, where the spatial domain is discretised into a triangular mesh with 415 nodes, and the temporal domain $t\in\left[0.05,\ 5\right]$ is discretised into 100 nodes with time interval $\delta t=0.05$. Then, 3500 sets of data are randomly generated; 3000 of them are used as training data, and the rest 500 groups are treated as test data.

\noindent \textbf{Results.} The quantitative error comparison of RO-NORM and baseline methods is presented in Table \ref{tbl1}.  It can be found that DeepONet has larger errors than other methods. The reason behind this phenomenon lies in the output of this task is the complex spatio-temporal function, and the DeepONet fails to deal with massive spatio-temporal locations due to the point-wise training manner. Although POD-DeepONet and PCA-Net are improved compared to DeepONet, there is still a large gap compared to RO-NORM. We suspect that this is due to an overall reduction of the spatio-temporal function, which will be analysed in the Discussion section 4.1. In this $a\left(\mathbf{x}\right)\rightarrow u\left(\mathbf{x},t\right)$ case, we expand the dimension in the middle layer to guarantee the formal feasibility of NORM. However, such expansion is random without a priori, which results in NORM achieving sub-optimal results, but with much lower predictive accuracy than RO-NORM and a large standard deviation. RO-NORM achieves the optimal prediction accuracy, improving the mean values of MME and relative $L_2$ error by 68.13$\%$ and 71.74$\%$, respectively, compared to NORM.

\begin{figure}[t]
\centering
\includegraphics[width=0.85\textwidth]{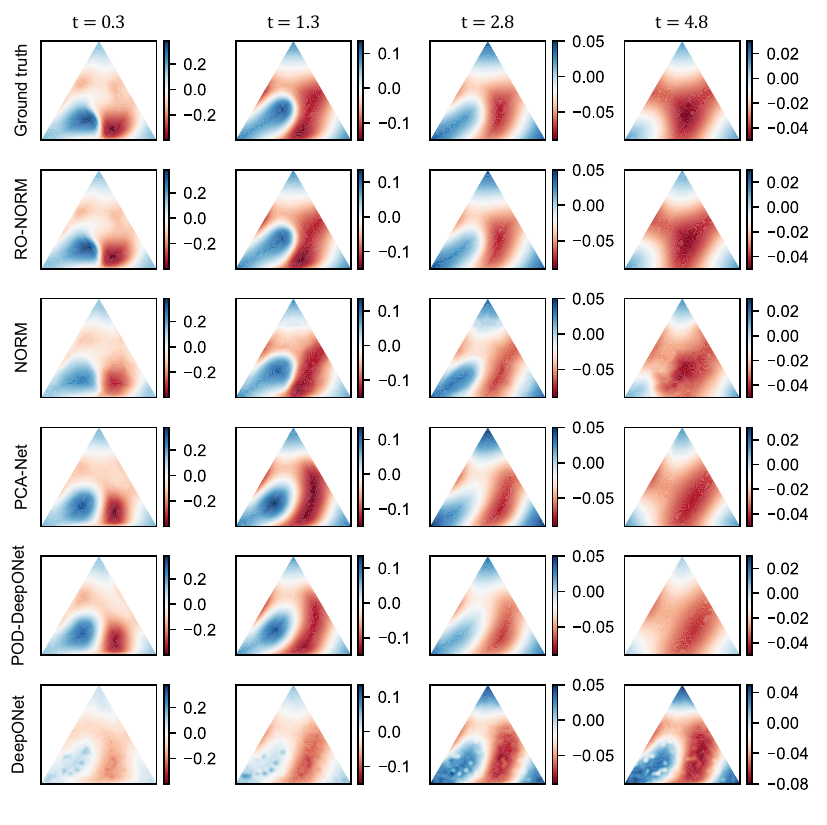}
\caption{\textcolor{black}{Comparison of solution accuracy between RO-NORM and baseline methods for the Burgers’ equations. Four representative time instants are chosen for display.}}
\label{Fig3}
\end{figure}

In particular, we select four snapshots of a representative test sample for illustration at $t=$ 0.3, 1.3, 2.8 and 4.8. \textcolor{black}{It can be seen in Fig. \ref{Fig3} that the predictions of RO-NORM possess excellent agreement with the ground truth. Besides, we observe that POD-DeepONet, PCA-Net and NORM capture the tendency of the field distribution but lose the local details. DeepONet fails to match the reference solution and shows a “spot” prediction. These phenomena are consistent with the results in Table \ref{tbl1} and further validate the superior solution accuracy of RO-NORM.} 

\subsection{2D Wave equations}
The wave equation is a second-order linear PDE, which describes the waves or standing wave fields arising in scenarios like acoustics, electromagnetism, and fluid dynamics. Here, we consider the undulation in a stretched elastic membrane caused by time-varying external perturbation, which can be formulated as a 2D wave equation:
\begin{equation}
    \frac{\partial^{2}f}{\partial t^{2}} = c^{2}\left( {\frac{\partial^{2}f}{\partial x^{2}} + \frac{\partial^{2}f}{\partial y^{2}}} \right) + v\left( {x,y,t} \right),~~~\mathbf{x} = \left( {x,y} \right) \in \lbrack 0,1\rbrack^{2},~t \in \lbrack 0,5\rbrack
\end{equation}
where $f\left(x,y,t\right)$ is the defection and $v\left(x,y,t\right)$ is the external perturbation referred to as the source term. $c$ denotes the wave propagation speed. 

\begin{figure}[t]
\centering
\includegraphics[width=0.85\textwidth]{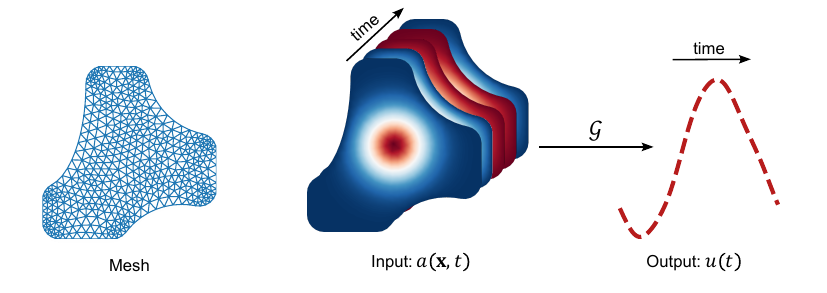}
\caption{Illustration of the 2D Wave equations.}
\label{Fig4}
\end{figure}

\begin{figure}[t]
    \centering
    \includegraphics[width=0.85\textwidth]{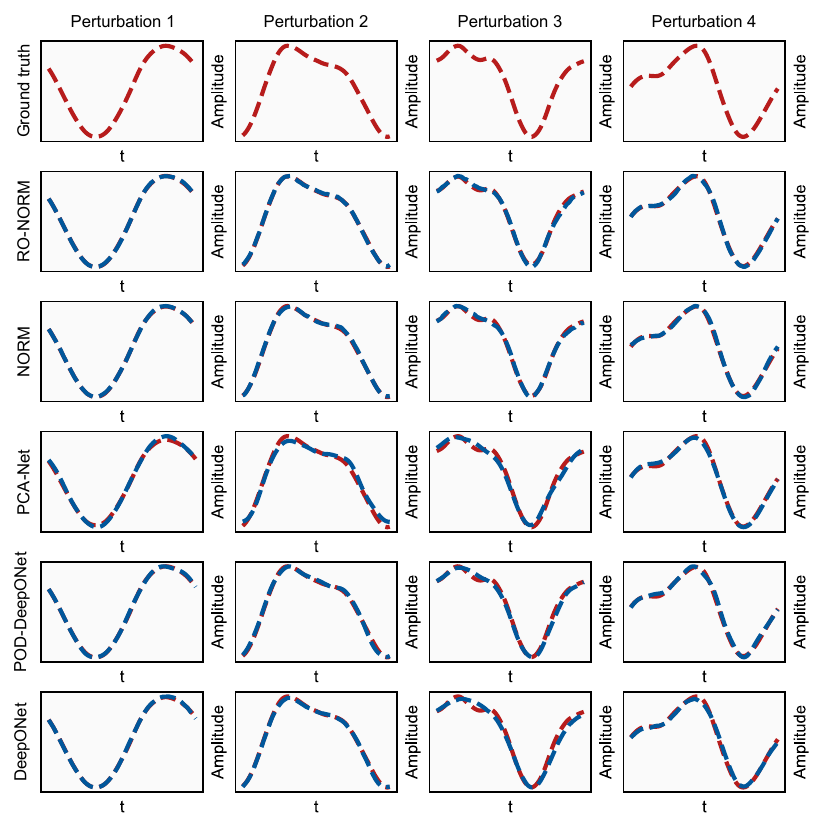}
    \caption{\textcolor{black}{Comparison of solution accuracy between RO-NORM and baseline methods for the wave equations. Four representative solution perturbations are chosen for display.}}
    \label{Fig5}
\end{figure}

\noindent\textbf{Problem setup.} We follow the source identification task in \cite{saha2021physics} and aim to approximate the operator mapping from observed long-term defection $a\left(\mathbf{x},t\right)=f\left(x,y,t\right)$ to the time-varying external perturbation $u\left(t\right)=v\left(t\right)$ at a fixed location: 
\begin{equation}
    \left. \mathcal{G}:\mathcal{~}a\left( {\mathbf{x},t} \right)\rightarrow u\left( t \right) \right.
\end{equation}

\noindent \textbf{Data generation.} As shown in Fig. \ref{Fig4}, the shape of the stretched elastic membrane is designed as a 2D irregular geometry. The time-varying perturbation $u\left(t\right)$ is fixed at the centre and is sampled randomly from a 1D Gaussian random field function: \textcolor{black}{\(GRF\  = \ GaussianRF(dim = 1,\ s = 100,\ alpha = 3.5,\ tau = 5,\ sigma = None,\ boundary = periodic)\).} Detailed implementation can be found in our source code. Then, we use COMSOL to calculate the long-term defection $a\left(\mathbf{x},t\right)$ with zero initial condition, homogeneous Dirichlet boundary condition, and $c^2 = 0.1$. The spatial domain is discretised into a triangular mesh with 506 nodes, and the temporal domain $t\in\left[0,\ 5\right]$ is discretised into 100 nodes with time interval $\delta t=0.05$. Finally, 2000 sets of data are randomly generated; 1500 of them are used as training data, and the rest 500 groups are treated as test data. 

\noindent \textbf{Results.} The statistical results of MME and relative $L_2$ error are given in Table \ref{tbl2}. \textcolor{black}{It can be observed that DeepONet no longer has a significant gap in accuracy with other methods as in the Burgers’ equation, and its accuracy is even slightly better than the POD-DeepONet. Because the output is the temporal function and the trunk net only needs to deal with time locations, which alleviates the modelling difficulty. In this $a\left(\mathbf{x},t\right)\rightarrow u\left(t\right)$ case, NORM achieves similar results to RO-NORM, slightly worse than RO-NORM. We consider the phenomena coming from the truth that the essence of shrinking domain dimension in the middle layer in NORM equals dimensionality reduction in RO-NORM, which can be regarded as feature extraction of the spatial domain dimension of the input function.}  

\begin{table}[t]
\caption{\textcolor{black}{Comparison of MME and $L_2$ relative error between RO-NORM and baseline methods on wave equation ($a\left( {\mathbf{x},t} \right)\rightarrow u\left(t\right)$).}}
\label{tbl2}
\begin{tabular}{lccccc|c}
\toprule
Metrics & DeepONet & POD-DeepONet & PCA-Net & NORM & RO-NORM & \textbf{\textit{Improvements}} \\
\midrule
MME & \textcolor{black}{0.016 (0.000)} & \textcolor{black}{0.017 (0.000)} & 0.027 (0.001) & \underline{0.011 (0.000)} & \textbf{0.010 (0.000)} & \textbf{\textit{9.09\% }}\\
$E_{L_2}(\%)$ & \textcolor{black}{6.213 (0.073)} & \textcolor{black}{6.511 (0.128)} & 10.969 (0.252) & \underline{3.875 (0.114)} & \textbf{3.635 (0.101)} & \textbf{\textit{6.19\%}} \\
\bottomrule
\end{tabular}
\end{table}

\textcolor{black}{Four representative solution perturbations predicted by RO-NORM and baselines, compared with the ground truth reference plotted in red dashed lines, are displayed in Fig. \ref{Fig5}. Herein, RO-NORM and NORM present outstanding goodness of fit with the ground truth, whereas PCA-Net and POD-DeepONet can simulate the general trend of the transient perturbations but lose the local details, such as perturbation 3. The identification of DeepONet is unstable, with good results on perturbations 1 and 2 but poor results on perturbations 3 and 4, corresponding to the larger standard deviations in Table \ref{tbl2}.}

\subsection{The temperature prediction of heat source layout }
As electronic components have become smaller and more complex, thermal management in systems engineering has been increasingly essential to ensure heat dissipation \cite{otaki2022thermal}. One practical approach is to optimise the positions of the heat-generating components, e.g. heat source layout optimisation, which can be described as the transient heat transfer equation:
\begin{equation}
    k\Delta T - \rho CT_{t} + \phi\left( {\mathbf{x},t} \right) = 0
\end{equation}
where k is the thermal conductivity, $ \phi\left(\mathbf{x},t\right)$ denotes the heat source, $\rho$ and C are the density and specific heat capacity of composites, respectively.

\noindent \textbf{Problem setup}. During the optimisation, there are massive iterative evaluations for temperature based on the corresponding heat source layout. Therefore, we tend to learn the mapping between the heat source layout $a\left(\mathbf{x}\right)$ and the time-dependent temperature field $T\left(\mathbf{x},t\right)$.
\begin{equation}
    \left. \mathcal{G}:\mathcal{~}a\left( \mathbf{x} \right)\rightarrow T\left( {\mathbf{x},t} \right) \right.
\end{equation}

\begin{figure}[t]
    \centering
    \includegraphics[width=0.85\textwidth]{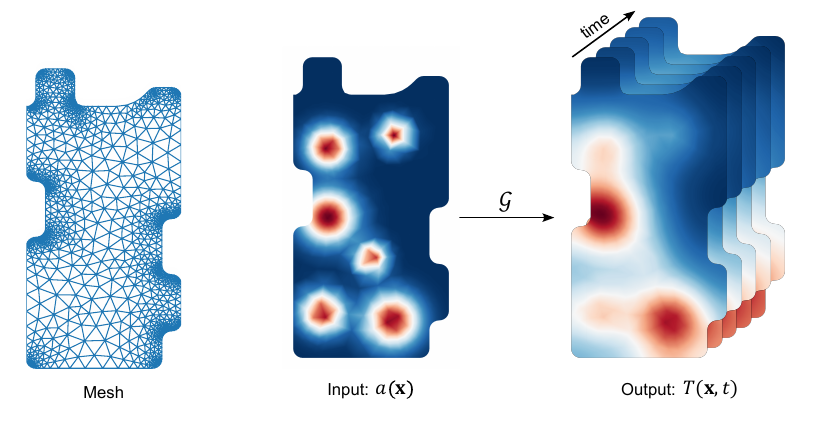}
    \caption{Illustration of the temperature prediction of heat source layout}
    \label{Fig6}
\end{figure}

\begin{figure}[htb]
    \centering
    \includegraphics[width=0.85\textwidth]{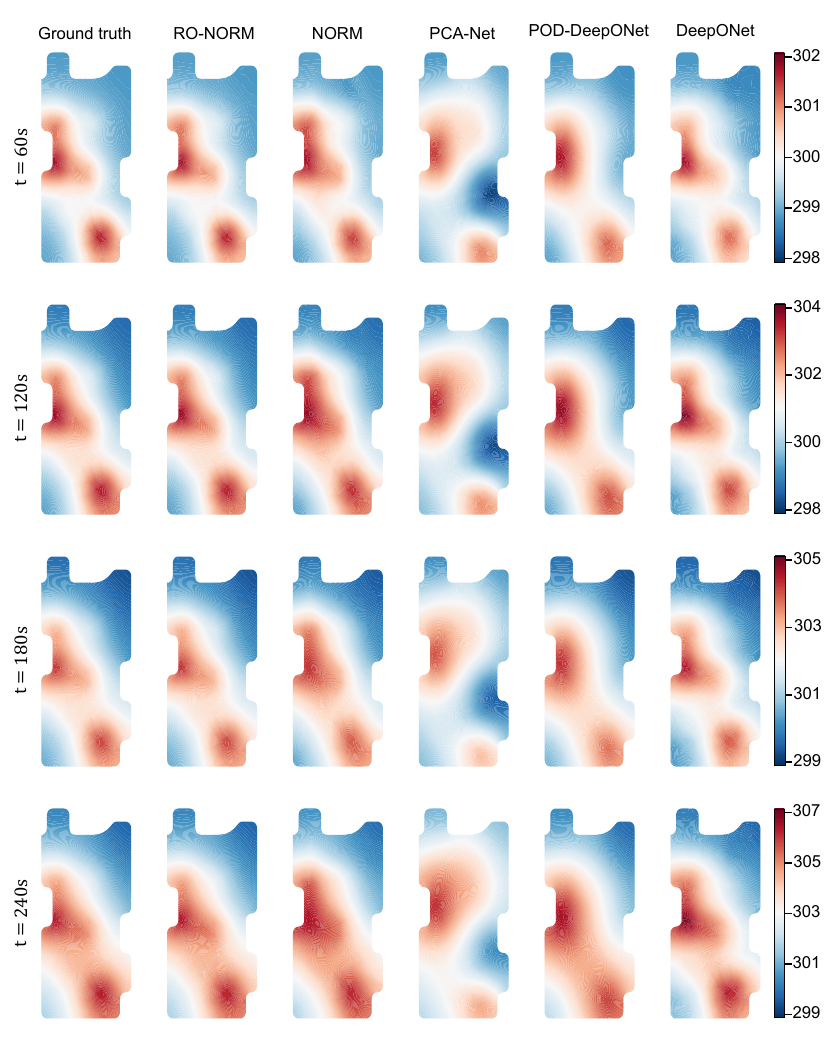}
    \caption{\textcolor{black}{Comparison of solution accuracy between RO-NORM and baseline methods for the heat source layout case. Four representative time instants are chosen for display.}}
    \label{Fig7}
\end{figure}

\noindent \textbf{Data generation.} Herein, we design a 2D irregular shape, like the phone cooling board, as shown in Fig. \ref{Fig6}. Assume that there are six circular heat sources with fixed sizes and fixed time-varying heating power. Then, we use a sequential and randomised approach to generate the locations of the six heat sources while ensuring they don’t overlap. The time-varying temperature field is calculated using COMSOL. The spatial domain is discretised into a triangular mesh with 1168 nodes, and the temporal domain $t\in\left[0,\ 300s\right]$ is discretised into 151 nodes with time interval $\delta t$=2s. Finally, 1200 sets of data are randomly generated; 1000 of them are used as training data, and the rest 200 groups are treated as test data. 

\noindent \textbf{Results.}
The quantitative comparison of MME and relative
\(L_{2}\) error between the proposed RO-NORM and other baselines is
shown in Table \ref{tbl3}. Additionally, we select four snapshots of a
representative test sample for illustration at \(t = 60s,\ 120s,180s\)
and \(240s\), as depicted in Fig. \ref{Fig7}.
\textcolor{black}{ We can observe that the predictions
of RO-NORM achieve a good match with the ground truth, corresponding to
the smallest MME and \(E_{L2}\) in Table \ref{tbl3}. The MME of DeepONet,
POD-DeepONet and NORM are similar, but \(E_{L2}\) of NORM is smaller,
which can explain the better predictive performance of NORM than
DeepONet and POD-DeepONet in Fig. \ref{Fig7}. Compared to RO-NORM, however, NORM still lose some detail. In this case, PCA-Net is less accurate than
other methods and has noticeable differences between predicted
temperature and reference fields.}

\begin{table}[htb]
\caption{\textcolor{black}{Comparison of MME and $L_2$ relative error between RO-NORM and baseline methods on heat source layout ($a\left(\mathbf{x}\right)\rightarrow T\left(\mathbf{x},t\right)$).}}
\label{tbl3}
\begin{tabular}{lccccc|c}
\toprule
Metrics & DeepONet & POD-DeepONet & PCA-Net & NORM & RO-NORM & \textbf{\textit{Improvements}} \\
\midrule
MME & \textcolor{black}{1.507 (0.019)} & \textcolor{black}{1.927 (0.005)} & 2.993 (0.048) & \underline{1.669 (0.084)} & \textbf{0.254 (0.013)} & \textbf{\textit{84.78\% }}\\
$E_{L_2}(\%)$ & \textcolor{black}{0.110 (0.001)} & \textcolor{black}{0.130 (0.000)} & 0.377 (0.006) & \underline{0.069 (0.008)} & \textbf{0.019 (0.001)} & \textbf{\textit{72.46\%}} \\
\bottomrule
\end{tabular}
\end{table}

\begin{figure}[t]
    \centering
    \includegraphics[width=0.85\textwidth]{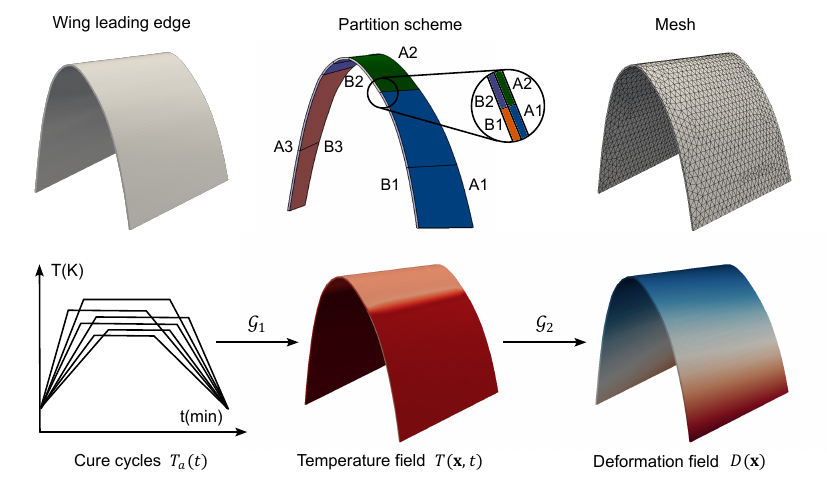}
    \caption{Illustration of composite workpiece temperature and deformation prediction. }
    \label{Fig8}
\end{figure}

\subsection{Composite workpiece temperature and deformation prediction}
Carbon-fibre-reinforced polymer (CFRP) composites have received much attention in aeronautical and astronautical structures \cite{barile2019mechanical}. Furthermore, the cure process is the necessary procedure for manufacturing CFRP composites, which converts the raw thermoset prepreg to the consolidated structure with the required mechanical properties and geometric shapes. Due to the chemical resin shrinkage, anisotropic thermal expansion, tool-part interaction and other factors, it is inevitable to cause the cure-induced deformation (CID) during the cure process, affecting the final quality of cured structures. To reduce deformation, Liu et al. \cite{liu2023active} propose a multi-zoned heating methodology . As shown in Fig. \ref{Fig8}, we consider a wing leading edge, where the internal and external surfaces of the workpiece are divided into six areas. A cure cycle is applied to each zone, which is a temperature function  $T_a\left(t\right)$ with respect to time, containing the ramp-up-dwell-cooling stages. To obtain the optimal cure cycles, we need to optimise the process parameters (heating rate, cooling rate, holding time and so on) based on the corresponding cure states, such as temperature and deformation, of which the key is to build predictive models. 

\subsubsection{Temperature prediction}

\noindent \textbf{Problem setup.} Cure cycles control the heat input of the composite part during the cure process, which is regarded as temperature boundary conditions. Therefore, we are interested in the mapping between the cure cycles $T_a(t)$ and the temperature field $T\left( \mathbf{x},t \right)$, as shown in Fig. \ref{Fig8}
\begin{equation}
    \left. \mathcal{G}_{1}:~T_a\left(t\right)\rightarrow T\left( \mathbf{x},t \right) \right.
\end{equation}

\noindent \textbf{Data generation.} Data generation. An aerospace grade composite material, AS4/8552 is considered in this case \cite{meng2023novel}. Each zone randomly samples a $300 \ min$ one-dwell cure cycle from the parameters ranges: heating stage time $t_1 \in[80 \ min, 90 \ min]$, dwell stage time $t_2 \in[140 \ min, 150 \ min]$, cooling stage time $t_3=350\ min-t_1-t_2$ and dwell temperature $T_1\in[160\ K,\ 190\ K]$. We conduct solid heat transfer coupled with cure kinetics simulations in COMSOL and consider Dirichlet boundary conditions for cure cycles. The spatial domain is discretised into a tetrahedral mesh with 2743 nodes, and the temporal domain $ t\in\left[0,\ 300\ min\right]$ is discretised into 151 nodes with time interval $\delta t=2\ min$. Finally, 600 sets of data are randomly generated; 500 of them are used as training data, and the rest 100 groups are treated as test data.

\begin{table}[t]
\caption{\textcolor{black}{Comparison of MME and $L_2$ relative error between RO-NORM and baseline methods on composite workpiece temperature prediction ($T_a\left(t\right)\rightarrow T\left( {\mathbf{x},t} \right)$).}}
\label{tbl4}
\begin{tabular}{lccccc|c}
\toprule
Metrics & DeepONet & POD-DeepONet & PCA-Net & NORM & RO-NORM & \textbf{\textit{Improvements}} \\
\midrule
MME & \textcolor{black}{40.973 (0.238)} & \textcolor{black}{7.351 (0.076} & \underline{6.711 (0.102)} & 32.308 (0.570) & \textbf{5.568 (0.083)} & \textbf{\textit{17.03\% }}\\
$E_{L_2}(\%)$ & \textcolor{black}{2.016 (0.003)} & \textcolor{black}{\underline{0.348 (0.003)}} & 0.365 (0.004) & 1.763 (0.011) & \textbf{0.257 (0.003) } & \textcolor{black}{\textbf{\textit{26.15\%}}} \\
\bottomrule
\end{tabular}
\end{table}

\begin{figure}[t]
    \centering
    \includegraphics[width=0.85\textwidth]{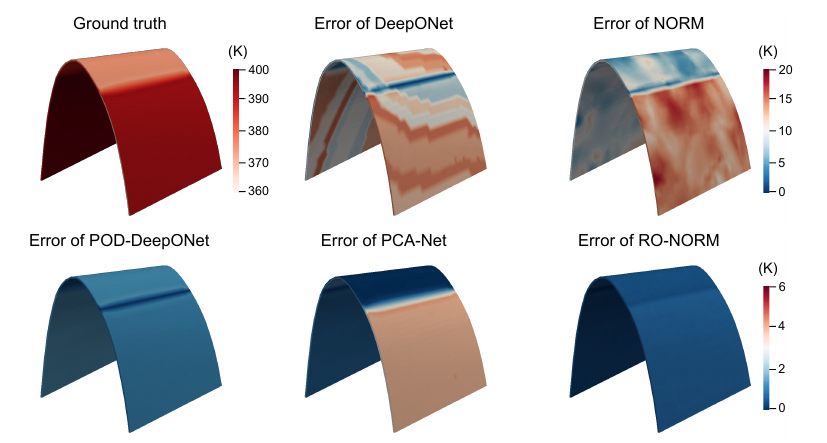}
    \caption{\textcolor{black}{Comparison of error fields between RO-NORM and baseline methods for the temperature prediction}.}
    \label{Fig9}
\end{figure}

\begin{figure}[t]
    \centering
    \includegraphics[width=0.85\textwidth]{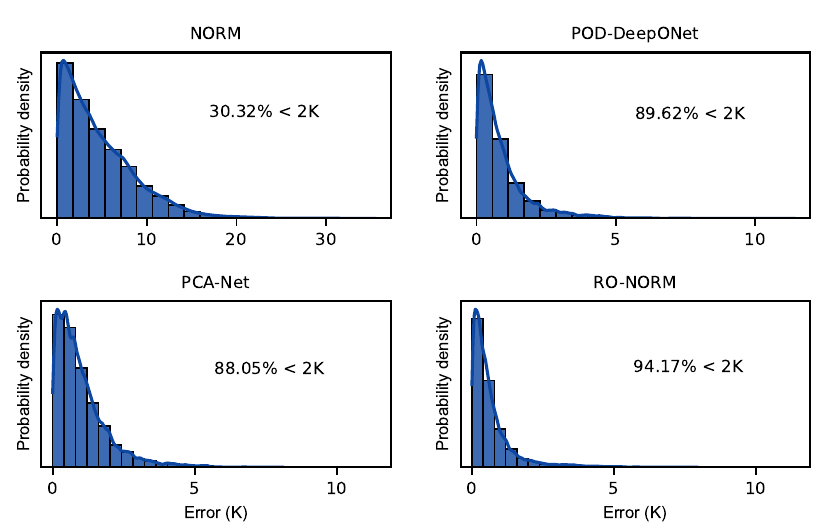}
    \caption{\textcolor{black}{The error probability density statistics of 600000 points in 100 test samples.}}
    \label{Fig10}
\end{figure}

\noindent \textbf{Results}. 
The results of MME and \(E_{L2}\) are listed in table \ref{tbl4}. RO-NORM obtain the lowest errors while NORM performs poorly, slightly
better than DeepONet. It reveals that expanding the domain dimension in
the middle layer cannot guarantee stable and expected performance. The
ground truth and the error fields of one representative moment of one
test sample are plotted in Fig. \ref{Fig9}. The prediction errors of DeepONet and
NORM are relatively large, and the distribution of errors is irregular,
where they don't show the characteristic of uniform temperature error in
each zone like PCA and POD-DeepONet. For further verification, we
randomly select 6000 spatio-temporal points, including 20 temporal and
300 spatial locations. The error probability density statistics of
600000 points in 100 test samples are given in Fig. \ref{Fig10}, where the
DeepONet is not plotted. \textcolor{black}{It can be observed that PCA-Net and RO-NORM
have smaller error ranges (i.e. x-axis ranges) compared to NORM and
POD-DeepONet, and have no points with errors of more than 10 K. Overall,
the error distribution of RO-NORM is closer to zero. The errors of more
than 94\% points of RO-NoRM are less than 2K, compared to 88.05\% for
PCA-Net and 89.62\% for POD-DeepONet, which is much higher than 30.32\%
for NORM.}

\subsubsection{Deformation prediction}

\noindent \textbf{Problem setup.} According to the idea of sequential coupling simulation, we tend to learn the operator mapping from the temperature field $T\left( {\mathbf{x},t} \right)$ to the deformation field $D\left( \mathbf{x} \right)$, as shown in Fig. \ref{Fig8}.
\begin{equation}
    \left. \mathcal{G}_{2}:~T\left( {\mathbf{x},t} \right)\rightarrow D\left( \mathbf{x} \right) \right.
\end{equation}

\noindent \textbf{Data generation.} We use the temperature field obtained from Section 3.4.1 as input and employ the CHILE constitutive relation to simulate the deformation in COMSOL, i.e. thermo-mechanical model. The discretisation of spatial and temporal domains and the division of training and test data remain the same as in Section 3.4.1.

\begin{figure}[t]
    \centering
    \includegraphics[width=0.85\textwidth]{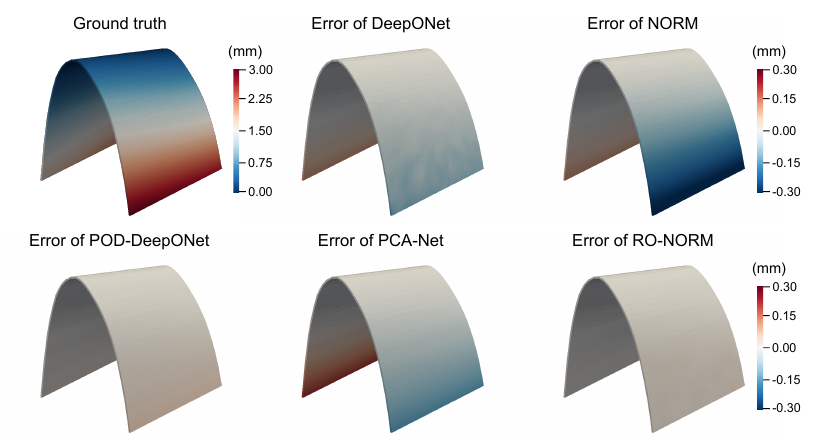}
    \caption{\textcolor{black}{Comparison of error fields between RO-NORM and baseline methods for the deformation prediction}.}
    \label{Fig11}
\end{figure}

\begin{figure}[t]
    \centering
    \includegraphics[width=0.85\textwidth]{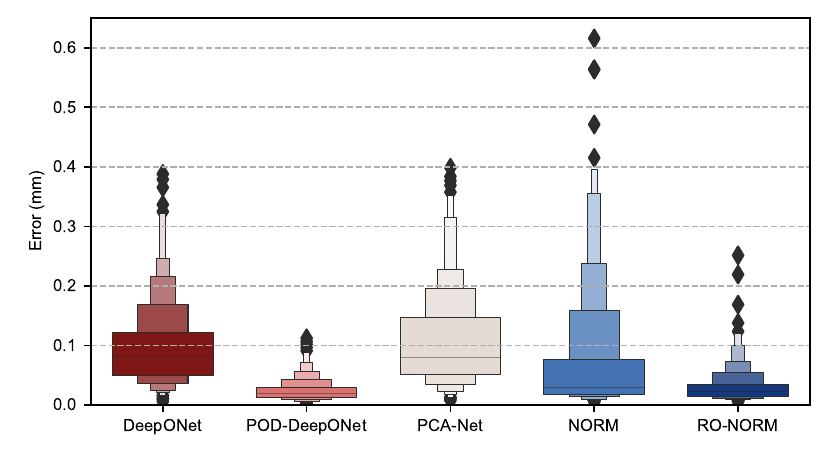}
    \caption{\textcolor{black}{The statistical analysis of the maximum prediction errors of all test samples for all methods of three repeated runs.}}
    \label{Fig12}
\end{figure}

\noindent \textbf{Results.} The results of MME and \(E_{L2}\) are listed in Table \ref{tbl5}. \textcolor{black}{Note that NORM has a larger standard deviation. The main reason is that the results of three repeat runs vary greatly, even with the same
training hyperparameters. This also demonstrates the instability of
shrinking dimensions for NORM when dealing with unequal-domain mappings.
As shown in Fig. \ref{Fig11}, NORM has a more significant error on a
representative test sample. Furthermore, we count the maximum prediction
error (ME) of all test samples for all methods of three repeated runs,
totalling 300 sets of results, as shown in Fig. \ref{Fig12}. PCA-Net and DeepONet
have similar performance, corresponding to the statistics results in
Table \ref{tbl5}. Although the ME of most of the NORM test samples is lower than
0.2mm, there is a significant portion of samples with larger maximum
errors, resulting in a large error span of zero to 0.7mm. POD-DeepONet
and RO-NORM far outperform other comparative methods, while POD-DeepONet
has lower MME and all test samples with a maximum prediction error of
less than 0.2mm.}

\begin{table}[htb]
\caption{\textcolor{black}{Comparison of MME and $L_2$ relative error between RO-NORM and baseline methods on composite workpiece deformation prediction ($T\left(\mathbf{x},t\right)\rightarrow D\left(\mathbf{x}\right)$).}}
\label{tbl5}
\begin{tabular}{lccccc|c}
\toprule
Metrics & DeepONet & POD-DeepONet & PCA-Net & NORM & RO-NORM & \textbf{\textit{Improvements}} \\
\midrule
MME & \textcolor{black}{0.097 (0.025)} & \textbf{\textcolor{black}{0.024 (0.001)}} & 0.107 (0.007) & 0.071 (0.059) & \underline{0.031 (0.002)} & \textcolor{black}{\textbf{\textit{-29.17\%}}} \\
$E_{L_2}(\%)$ & \textcolor{black}{2.346 (0.578)} & \underline{\textcolor{black}{0.570 (0.034)}} & 2.684 (0.235) & 1.734 (1.641) & \textbf{0.549 (0.050)} & \textcolor{black}{\textbf{\textit{3.68\% }}}\\
\bottomrule
\end{tabular}
\end{table}

\subsection{Blood flow dynamics prediction}
In the last two decades, there has been a globally increasing in vascular diseases, which is the leading cause of death \cite{mendis2011global}. Therefore, it is desirable to study the characteristics and regularities of the movement of blood, i.e. blood flow dynamics, that can promote the diagnosis and treatment of vascular diseases. In addition, with the development of computational fluid dynamics (CFD), CFD modelling has been widely used to simulate blood flow by numerically solving the Navier-Stokes equations \cite{caballero2013review}. Despite its outstanding predictive performance, CFD modelling's high computing cost and long processing time have limited its use in clinical practice in time-sensitive areas such as pre-operative planning and serial monitoring \cite{kissas2020machine}. In this section, therefore, we aim to explore the potential of data-driven neural operator models for the surrogate modelling of haemodynamic CFD.

\begin{figure}[htb]
    \centering
    \includegraphics[width=0.85\textwidth]{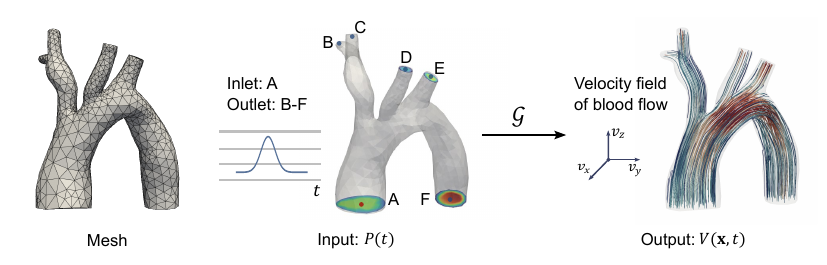}
    \caption{Illustration of blood flow dynamics prediction.}
    \label{Fig13}
\end{figure}

\noindent \textbf{Problem setup}. Herein, we focus on the aorta and consider a similar scenario as described in \cite{maul2023transient}, which inputs the time-varying pressure and velocity at the inlet/outlets $P\left(t \right)$ and evaluates the velocity field $V\left( \mathbf{x},t \right)$ composed of velocity components in three directions, as displayed in Fig. \ref{Fig13}. 

\begin{equation}
    \left. \mathcal{G}:\mathcal{~} P\left(t \right)\rightarrow V\left( \mathbf{x},t \right) \right.
\end{equation}

\noindent \textbf{Data generation.} The details of data generation are the same as those in NORM \cite{Chen2023LearningNO}. A total of 500 groups of labelled data are generated. 400 of them are used as training data, and the rest 100 are used as test data.

\begin{figure}[htb]
    \centering
    \includegraphics[width=0.85\textwidth]{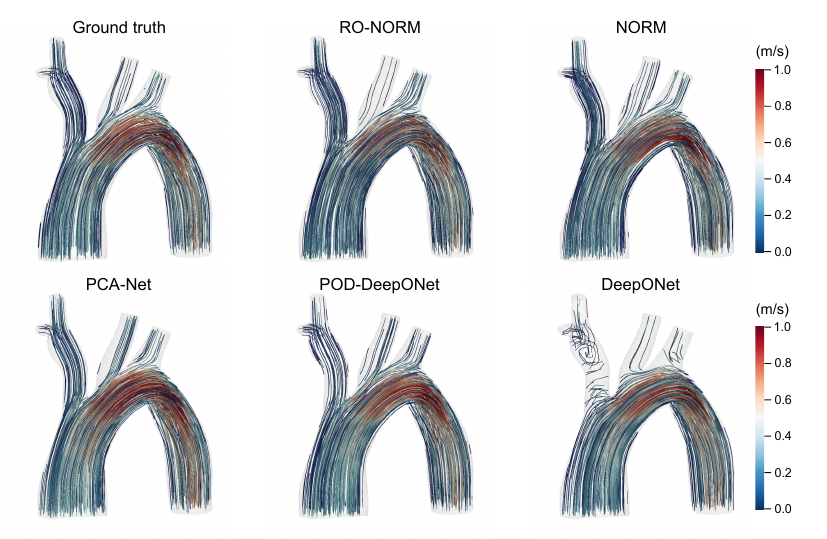}
    \caption{\textcolor{black}{Comparison of velocity streamlines between RO-NORM and baseline methods}.}
    \label{Fig14}
\end{figure}

\noindent \textbf{Results.} The MME and \(E_{L2}\) results of this case are
summarised in Table \ref{tbl6}. The visualisation of the velocity streamlines
(snapshots at a representative time) against baseline methods is
provided in Fig.14. It is worth pointing out that this case faces the
challenge of unbalanced node values, i.e. the velocity of most nodes is
close to zero due to the no-slip boundary condition. Notably, a slight
prediction bias at nodes \(v \rightarrow 0\) would cause a significant
relative error. \textcolor{black}{Therefore, we can see that the \(E_{L_{2}}\) of RO-NORM is improved by 27.35\% compared to NORM, but the MME is increased from
0.076 to 0.087. This suggests that RO-NORM may pay more attention to
points where the velocity is close to zero. Although the outputs are
spatio-temporal, the \(E_{L2}\) of POD-DeepONet is also below 5\%
because of the simple evolution mode during the systolic and diastolic
phases. Fig.14 demonstrates the remarkable approximation capability of
RO-NORM, NORM and POD-DeepONet. PCA-Net has a moderate performance and
learns the general trend of velocity distribution but loses some of the
predictive accuracy of the node. As shown in Fig. \ref{Fig14} and Table \ref{tbl6}, the
point-wise training manner of DeepONet leads to a significant \(L_{2}\)
relative error 48.503\%. Especially, DeepONet fails to capture the local
details of streamlines outlets.}

\begin{table}[htb]
\caption{\textcolor{black}{Comparison of MME and $L_2$ relative error between RO-NORM and baseline methods on blood flow dynamics prediction ($P\left(t \right)\rightarrow V\left( \mathbf{x},t \right)$).}}
\label{tbl6}
\begin{tabular}{lccccc|c}
\toprule
Metrics & DeepONet & POD-DeepONet & PCA-Net & NORM & RO-NORM & \textbf{\textit{Improvements}} \\
\midrule
MME & \textcolor{black}{0.715 (0.006)} & \textcolor{black}{0.112 (0.001)} & 0.175 (0.006) & \textbf{\textcolor{black}{0.076 (0.001)}} & \underline{0.087 (0.001)} & \textcolor{black}{\textbf{\textit{-13.16\%}}} \\
$E_{L_2}(\%)$ & \textcolor{black}{48.503 (0.006)} & \textcolor{black}{3.880 (0.049)} & 10.873 (0.728) & \underline{\textcolor{black}{3.536 (0.039)})} & \textbf{2.569 (0.016)} & \textcolor{black}{\textbf{\textit{27.35\%}}} \\
\bottomrule
\end{tabular}
\end{table}

\section{Discussion}
\subsection{Comparison between RO-NORM and POD-DeepONet/ PCA-Net}
\begin{figure}[htb]
    \centering
    \includegraphics[width=0.85\textwidth]{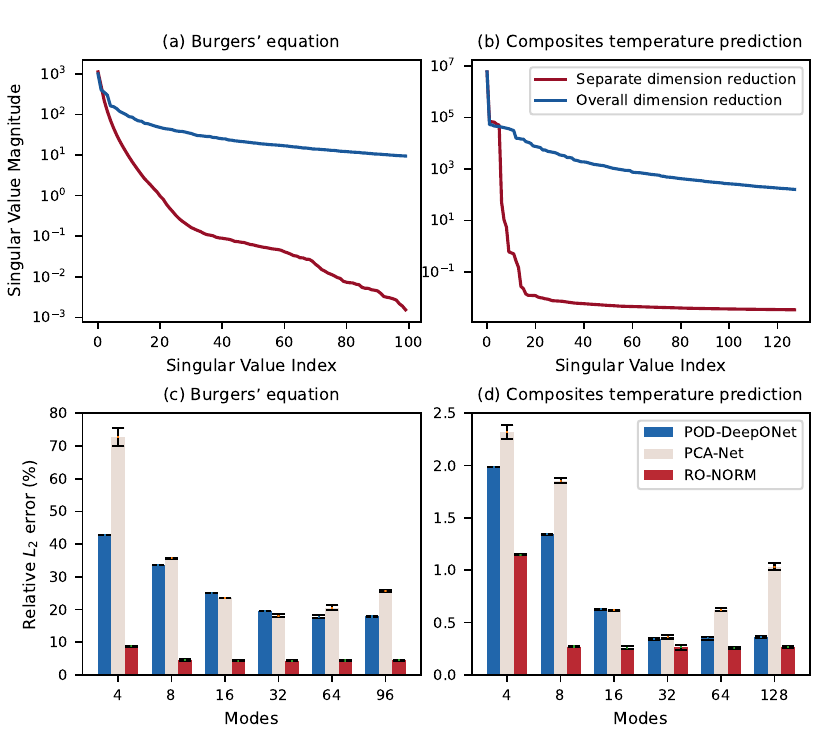}
    \caption{\textcolor{black}{The decay of singular values and the relative $L_2$ errors of three methods under different numbers of truncated modes. }.}
    \label{Fig15}
\end{figure}

As shown in Section 3, RO-NORM has obtained more accurate results than POD-DeepONet and PCA-Net in most cases when the number of truncated modes remains the same. To explore the hyperparameter impact on the three methods, we conduct quantitative performance comparisons on Burgers’ equation and composites temperature prediction, where the outputs are spatio-temporal functions. \textcolor{black}{The first row of Fig. \ref{Fig15} shows the decay of singular values for two cases with separate and overall dimension reduction, and the second row gives the relative $L_2$ errors of three methods under different numbers of truncated modes with three repeated runs. } 
 
\textcolor{black}{The decay rate of Kolmogorov $n$-width can quantify the optimal linear subspace \cite{fries2022lasdi}. If the Kolmogorov $n$-width decays fast, the dimensionality reduction can provide an effective subspace that can accurately approximate a true solution. The rate of singular value decay can well indicate the decay rate in Kolmogorov $n$-width. For the spatio-temporal function, the separate dimension reduction used in RO-NORM has a faster decay than overall dimension reduction, as shown in Fig. \ref{Fig15} (a)-(b). This explains why RO-NORM achieves good results when the number of modes is very small. In addition, as the number of modes increases, the errors of POD-DeepONet and NORM start to decrease and stabilise beyond a certain level, as shown in Fig. \ref{Fig15} (c)-(d). Similar conclusions are also mentioned in Ref \cite{jha2024residual}. The relative $L_2$ errors of PCA-Net show a tendency to decrease and then increase, which may be because adding more truncated modes increases the difficulty of FC-NN in learning the mapping between two higher-dimensional subspaces. This comparison highlights that the RO-NORM exhibits superior robustness for the number of truncated modes, making hyperparameter decisions easier.} 

\subsection{Comparison between RO-NORM and traditional neural networks}
The verification goal of this section includes two aspects: (1) to verify whether the dimensionality reduction using the separation of variables makes the transformed same-domain mapping easy to model, (2) to demonstrate the merit of NORM compared with traditional neural networks. \textcolor{black}{We choose wave equations
\(a\left( \mathbf{x},t \right) \rightarrow u\left(t\right)\) and blood flow
dynamics prediction \(P\left(t\right) \rightarrow V\left(\mathbf{x},t\right)\) for validation
and design two baseline models with two traditional neural networks to
model the transformed mapping \(w\left(t\right) \rightarrow u\left(t\right)\)
(\(P\left(t\right) \rightarrow w\left(t\right)\)) as comparisons:}\textcolor{black}{
\begin{itemize}
\item
  RO-FC-NN: Use the fully connected neural network (FC-NN), a classical
  architecture, to replace the NORM in RO-NORM. The FC-NN in this
  section has four hidden layers, each including 256 neurons. Other
  parameter settings are kept the same with RO-NORM.
\item
  RO-U-net: Use the CNN-based U-net to replace the NORM in RO-NORM.
  U-Net, a U-shaped convolutional neural network, excels in medical
  image segmentation \cite{ronneberger2015u} by merging downsampling and upsampling
  layers for precise detail, offering localized accuracy and efficiency.
  The U-net used in this section includes four downsampling layers and
  four upsampling layers. Other parameter settings are kept the same
  with RO-NORM.
\end{itemize}
}\textcolor{black}{
The MME and \(E_{L2}\) results are summarised in Table \ref{tbl7} and Table \ref{tbl8},
where the results of PCA-Net and RO-NORM are copied from Section 3 for
ease of reference. Since the transformed mappings belong to
sequence-to-sequence mapping, it is reasonable for RO-U-net to have
better results than RO-FC-NN. Besides, it can be observed that RO-U-net
achieves more accurate predictions than PCA-Net in two cases. Even
RO-FC-NN reaches a similar level with PCA-Net in the blood flow case.
This comparison demonstrates the effectiveness of dimensionality
reduction using the separation of variables to reduce the difficulty of
modelling. Compared to the vector mapping simplified by PCA-Net, the
mapping simplified by RO-NORM maintains the infinite property of the
operator. Despite the appropriate architecture, such as U-net, leading
to good performance, there remains a significant gap between traditional
neural networks and NORM modelling, reflecting the gap between the two
in terms of operator learning.
}

\begin{table}[htb]
\caption{Comparison of MME and $L_2$ relative error on wave equation ($a\left( {\mathbf{x},t} \right)\rightarrow u\left(t\right)$). Note that FC-NN and U-net are used as the same-domain approximator in RO-NORM..}
\label{tbl7}
\begin{tabular}{lccccc}
\toprule
Metrics  & PCA-Net & RO-FC-NN & RO-U-net & RO-NORM \\
\midrule
MME  & 0.027 (0.001) & 0.144 (0.000) & 0.017 (0.001) & \textbf{0.010 (0.000)}  \\
$E_{L_2}(\%)$ & 10.969 (0.252) & 15.340 (0.054) & 5.828 (0.746) & \textbf{3.635 (0.101)} \\
\bottomrule
\end{tabular}
\end{table}

\begin{table}[htb]
\caption{Comparison of MME and $L_2$ relative error on blood flow dynamics prediction ($P\left(t\right)\rightarrow V\left( {\mathbf{x},t} \right)$). Note that FC-NN and U-net are used as the same-domain approximator in RO-NORM.}
\label{tbl8}
\centering
\begin{tabular}{lccccc}
\toprule
Metrics  & PCA-Net & RO-FC-NN & RO-U-net & RO-NORM \\
\midrule
MME  & 0.175 (0.006) & 0.246 (0.002) & 0.130 (0.003) & \textbf{0.087 (0.001)}  \\
$E_{L_2}(\%)$  & 10.873 (0.728) & 11.208 (0.144) & 5.602 (0.240) & \textbf{2.569 (0.016)} \\
\bottomrule
\end{tabular}
\end{table}

\subsection{Training efficiency comparison between RO-NORM and NORM}

To further assess the proposed RO-NORM, we conduct comparison
experiments to investigate the training efficiency and prediction
performance of RO-NORM and NORM. To distinguish increase-domain mapping
and decrease-domain mapping, therefore, we select wave equations
\(a(x,t) \rightarrow u(t)\) and heat source layout
\(a(x) \rightarrow u(x,t)\) as representative cases and fix the total
computational budget for training, especially both the RO-NORM and NORM
are trained with 100 iterations and the same structure setting, such as
\(\mathcal{P}\), \(\mathcal{Q}\), the number of truncated modes and the
number of L-layers. It can be seen from Table \ref{tbl9} that the MME and
\(E_{L2}\) of NORM are slightly worse than those of RO-NORM in the
decrease-domain mapping, which means shrinking the domain dimension in
the middle layer in NORM would achieve a similar effect as dimension
reduction in RO-NORM. Even so, there is a more than five times
difference in training time. In the increase-domain mapping, because
NORM has to expand the domain dimension in the middle layer to match the
output and then implement the kernel integration operator separately for
the time and space dimensions in each subsequent L-layer, NORM achieves
approximately ten times larger error and twenty times longer training
time. The comparative study has demonstrated the
RO-NORM\textquotesingle s fast training and higher accuracy,
highlighting its potential for practical applications.
    
\begin{table}[htb]
\caption{Comparison of predictive accuracy of RO-NORM and NORM with the same amount of training budget.}
\label{tbl9}
\centering

\begin{tabular}{lcccccc}
\toprule
Case & Iterations & Information & RO-NORM & NORM & Difference & Device name \\
\midrule
\multirow{4}{*}{\makecell[l]{Wave equations  \\  $a\left( {\mathbf{x},t} \right)\rightarrow u\left(t\right)$}} 
& \multirow{4}{*}{100} & Parameters & 23793 & 23874 & \textbf{1.01x }
& \multirow{4}{*}{\makecell[c] {NVIDIA \\ GeForce \\ GTX 1660 SUPER}} \\
& & Wall-clock time (s) & 26.139 & 147.087 & \textbf{5.63x} & \\
& & MME & 0.014 & 0.017 & \textbf{1.21x} & \\
& & $E_{L_2}(\%)$ & 5.239 & 6.218 & \textbf{1.19x} & \\
\midrule
\multirow{4}{*}{\makecell[l]{Heat source layout \\ $a\left(\mathbf{x}\right)\rightarrow u\left(\mathbf{x},t\right)$}} 
& \multirow{4}{*}{100} & Parameters & 133264 & 2623985 & \textbf{19.69x} 
& \multirow{4}{*}{\makecell[c] {NVIDIA \\ GeForce \\ GTX 1660 SUPER}} \\
& & Wall-clock time (s) & 102.077 & 2379.363 & \textbf{23.31x} & \\
& & MME & 0.616 & 4.757 & \textbf{7.72x} & \\
& & $E_{L_2}(\%)$ & 0.049 & 0.504 & \textbf{10.29x} & \\
\bottomrule
\end{tabular}
\end{table}

\begin{table}[htb]
\caption{Comparison of predictive accuracy of RO-NORM with data-dependent and data-independent bases.}
\label{tbl10}
\centering
\begin{tabular}{lcccc}
\toprule
Cases & Metrics & POD & LBO/Fourier \\
\midrule
\multirow{2}{*}{\makecell[l]{Burgers’ equations \\ $a\left(\mathbf{x}\right)\rightarrow u\left(\mathbf{x},t\right)$}} & MME & \textbf{0.065(0.001)} & 0.230(0.001) \\
 & $E_{L_2}(\%)$ & \textbf{4.356(0.066)} & 11.299(0.041) \\
\midrule
\multirow{2}{*}{\makecell[l]{Wave equations \\ $a\left( {\mathbf{x},t} \right)\rightarrow u\left(t\right)$}} & MME & \textbf{0.010(0.000)} & 0.012(0.000) \\
 & $E_{L_2}(\%)$ & \textbf{3.635(0.101)} & 4.229(0.105) \\
\midrule
\multirow{2}{*}{\makecell[l]{Composites temperature prediction \\ $a\left(t\right)\rightarrow u\left( {\mathbf{x},t} \right)$}} & MME & \textbf{5.493(0.047)}  & 29.501(0.054) \\
& $E_{L_2}(\%)$ & \textbf{0.258(0.002)} & 1.746(0.001) \\
\midrule
\multirow{2}{*}{\makecell[l]{Composites deformation prediction \\ $a\left(\mathbf{x},t\right)\rightarrow u\left(\mathbf{x}\right)$ }}& MME & 0.031(0.002) & \textbf{0.023(0.002)} \\
& $E_{L_2}(\%)$ & 0.549(0.050) & \textbf{0.373(0.015)} \\
\bottomrule
\end{tabular}
\end{table}

\subsection{Comparison between data-dependent and data-independent bases}

\begin{figure}[htb]
    \centering
    \includegraphics[width=0.85\textwidth]{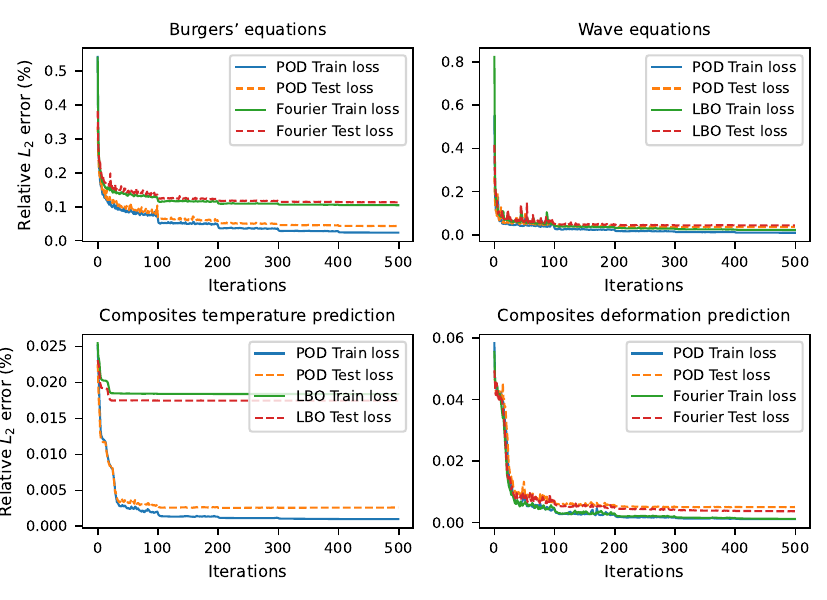}
    \caption{\textcolor{black}{The convergent results for RO-NORM with data-driven and intrinsic basis.}}
    \label{Fig16}
\end{figure}

In Section 3, we applied POD-derived in RO-NORM bases to reduce the temporal or spatial dimensions in the input or output. In addition to these data-dependent bases, intrinsic bases such as Laplace eigenfunctions, which are data-independent and reflect the inherent properties, can also be used. Comparative experiments are conducted on four unequal-domain mappings of PL-STP to investigate the predictive accuracy of RO-NORM under data-dependent and data-independent bases, where Fourier basis and LBO basis are adopted as intrinsic bases in temporal and spatial domains, respectively. The error statistics and convergence results for the four cases are presented in Table \ref{tbl10} and Fig. \ref{Fig16}. The results can still be analysed according to increase-domain and decrease-domain mappings. In the domain-increase mappings, the POD basis outperforms the intrinsic basis in performance. This is because the reconstruction accuracy of the intrinsic basis is much lower than that of the data-dependent basis for the same number of truncated modes. For example, the LBO basis is solely related to the intrinsic structure of the manifold and can represent any function defined on it. However, to achieve favourable reconstruction accuracy, a large number of truncations are required due to the unknown regularity of the function's distribution. On the other hand, in the decrease-domain mapping, it can be concluded that the performance of the POD and the intrinsic bases is similar, and even the intrinsic basis performs slightly better. It is worth noting that this scenario no longer relies on the reconstruction ability of the basis but the representation ability of the basis for reducing the dimension. These two abilities are not contradictory since a set of bases, especially intrinsic bases, can generally represent different original functions as different coefficient vectors but introduce significant truncation errors.

\subsection{Comparison between online and offline training}

As previously stated, it is essential for RO-NORM to reconstruct the spatio-temporal function of the output with the weight coefficient function predicted by the model and the pre-computed basis in domain increase cases. Depending on whether the reconstruction is considered in the training process, the training pattern can be divided into offline and online reconstruction. In the above sections, we use the online reconstruction way. The comparison results of the four increase-domain mappings are tabulated in Table \ref{tbl11}. From the error results, it is clear that online reconstruction consistently results in smaller errors than offline reconstruction, indicating that non-end-to-end offline reconstruction may be at risk of failure due to the direct learning bias of the weight coefficient function.

\begin{table}[htb]
\caption{Comparison of predictive accuracy of RO-NORM with offline and online reconstruction.}
\label{tbl11}
\centering
\begin{tabular}{lcccc}
\toprule
Cases & Metrics & Offline & Online \\
\midrule
\multirow{2}{*}{\makecell[l]{Burgers’ equations \\ $a\left(\mathbf{x}\right)\rightarrow u\left(\mathbf{x},t\right)$}} & MME & 0.158(0.005) & \textbf{0.065(0.001)} \\
 & $E_{L_2}(\%)$ & 17.282(0.147) & \textbf{4.356(0.066)} \\
\midrule
\multirow{2}{*}{\makecell[l]{Heat source layout \\$a\left(\mathbf{x}\right)\rightarrow u\left(\mathbf{x},t\right)$}} & MME & 0.388(0.027) & \textbf{0.254(0.013)} \\
 & $E_{L_2}(\%)$ & 0.031(0.002) & \textbf{0.019(0.001)} \\
\midrule
\multirow{2}{*}{\makecell[l]{Composites temperature prediction \\ $a\left(t\right)\rightarrow u\left( {\mathbf{x},t} \right)$}} & MME & 6.055(0.095) & \textbf{5.568(0.083)} \\
 & $E_{L_2}(\%)$ & 0.287(0.004) & \textbf{0.257(0.003)} \\
\midrule
\multirow{2}{*}{\makecell[l]{Blood flow dynamics \\ $a\left(t\right)\rightarrow u\left( {\mathbf{x},t} \right)$ }} & MME & 0.171(0.002) & \textbf{0.087(0.001)} \\
 & $E_{L_2}(\%)$ & 7.534(0.063) & \textbf{2.569(0.016)} \\
\bottomrule
\end{tabular}
\end{table}

\section{Conclusion}

This paper presents a general reduced-order neural operator called
RO-NORM for unequal-domain mappings with complex spatial domains in
PL-STP, where the input and output functions involve spatio-temporal and
spatial (or temporal) functions:
\(a\left(\mathbf{x}\right) \rightarrow u\left(\mathbf{x},t\right)\),
\(a\left(\mathbf{x},t\right) \rightarrow u\left(\mathbf{x}\right)\),
\(a\left(t\right) \rightarrow u\left(\mathbf{x},t\right)\),
\(a\left(\mathbf{x},t\right) \rightarrow u\left(\mathbf{x}\right)\). \textcolor{black}{The framework of RO-NORM includes an unequal-domain encoder/decoder and a same-domain
approximator NORM. Based on the separation of variables in classical
modal decomposition, the unequal-domain encoder/decoder transforms the
unequal-domain mapping into the same-domain mappings so that the
approximator NORM can be applied directly without performing expansion
(for increase-domain mapping) or shrinkage (for decrease-domain mapping)
at the input or middle layer to match the input and output domains.
Comparison experiments on six benchmark cases demonstrate that RO-NORM
significantly enhances the accuracy and stability of neural operators.}
Compared to NORM, RO-NORM improves the predictive accuracy while significantly increasing the training efficiency. Compared to PCA-Net, the accuracy improvement of RO-NORM is more apparent, and the variable-separated dimensionality reduction adopted by RO-NORM not only reduces the difficulty of modelling but also enables RO-NORM to show superior robustness for the number of truncated bases. Finally, we conduct a series of characteristic analyses, including the choice of basis and the difference between online and offline training, which will help push RO-NORM forward to practical applications.

\section{Appendix}
The details on selecting hyperparameters are provided in the Table \ref{tbl12}.

\begin{table}[pos=htbp]
\caption{\textcolor{black}{The details on the selection of hyperparameters. The \textit{Branch} and \textit{Trunk} mean the structures of hidden layers. The \textit{Lr} means learning rate and \textit{StepLR} is the learning rate decay strategy. The \textit{truncated modes} of POD-DeepONet, PCA-Net and RO-NORM keep the same. \textit{Lmodes} indicates the truncted number of Laplacian eigenfunctions used in L-layers, where the \textit{Lmodes1} and \textit{Lmodes2} correspond to LBO bases and Fourier bases respectively. The \textit{width} means the lifted channels after lifting layer \(\mathcal{P}\).}} 
\label{tbl12}
\centering
\begin{tabular}{llllllll}
\toprule
Methods	& Setting	&Burgers	&Wave	&Layout	&Temperature	&Deformation	&BloodFlow \\
\midrule
\multirow{3}{*}{Data size} & Traindata	&3000	&1500	&1000	&500	&500	&400\\
& Testdata	&500	&500	&200	&100	&100	&100\\
& Batchsize	&50	&50	&50	&50	&50	&10\\
\midrule
\multirow{5}{*}{DeepONet} &Branch	&256*256*256	&256*256*256	&256*256*256	&256*256*256	&256*256*256	&256*256*256\\
& Trunk	&256*256*256	&256*256*256	&256*256*256	&256*256*256	&256*256*256	&256*256*256\\
& Epoches	&10000	&10000	&5000	&5000	&5000	&5000\\
&Lr	&1e-3	&1e-3	&1e-3	&1e-3	&1e-3	&1e-3\\
&StepLR	&0.5(1000)	&0.5(1000)	&0.5(1000)	&0.5(1000)	&0.5(1000)	&0.5(1000)\\

\midrule
\multirow{5}{*}{POD-DeepONet} & \makecell[l]{Truncated\\modes}	&32	&16	&8	&32	&64	&64\\
& Branch	&256*256*256	&256*256*256	&256*256*256	&256*256*256	&256*256*256	&256*256*256\\
& Epochs	&10000	&10000	&5000	&5000	&5000	&10000\\
& Lr	& 1e-3	& 1e-3	& 1e-3	& 1e-3	& 1e-3	& 1e-3\\
& StepLR	& 0.5(1000)	& 0.5(1000)	& 0.5(1000)	& 0.5(1000)	& 0.5(1000)	& 0.5(1000)\\

\midrule
\multirow{5}{*}{PCA-Net} & \makecell[l]{Truncated\\modes}	& 32	& 16	& 8	& 32	& 64	& 64\\
& Net	& \makecell[l]{256*256*\\256*256}	& \makecell[l]{256*256*\\256*256}	& \makecell[l]{256*256*\\256*256}	& \makecell[l]{256*256*\\256*256}	& \makecell[l]{256*256*\\256*256}	& \makecell[l]{256*256*\\256*256}\\
& Epochs	&10000	&10000	&10000	&20000	&20000	&20000\\
& Lr	&0.0001	&0.0001	&0.0001	&0.0001	&0.0001	&0.0001\\
  
\midrule
\multirow{7}{*}{NORM} &Lmodes1	&128	&128	&128	&64	&128	&64\\
& Lmodes2	&16	&16	&16	&16	&16	&16\\
& L-layers	&4 Layers	&4 Layers	&4 Layers	&4 Layers	&4 Layers	&4 Layers\\
& width	&16	&16	&16	&16	&32	&16\\
& Epochs	&500	&500	&500	&200	&300	&500\\
& Lr	&0.1	&0.01	&0.1	&0.05	&0.01	&0.001\\
& StepLR	&0.1(100)	&0.5(100)	&0.1(100)	&0.1(100)	&0.5(100)	&0.1(100)\\

\midrule
\multirow{7}{*}{RO-NORM} &\makecell[l]{Truncated\\modes}	&32	&16	&8	&32	&64	&64\\
&Lmodes	&128 &32	&128	&16	&128	&16\\
&L-layers	&4 Layers	&4 Layers	&4 Layers	&4 Layers	&4 Layers	&4 Layers\\
&Width	&64	&16	&16	&32	&32	&64\\
&Epochs	&500	&500	&500	&500	&500	&500\\
&Lr	&0.01	&0.01	&0.01	&0.01	&0.01	&0.001\\
&StepLR	&0.5(100)	&0.5(100)	&0.5(100)	&0.5(100)	&0.5(100)	&0.1(100)\\

\bottomrule
\end{tabular}
\end{table}





\printcredits

\section*{Data and code availability}
The datasets of all six case studies and the source code of the RO-NORM are available at \href{https://github.com/qingluM/RO-NORM}{https://github.com/qingluM/RO-NORM}.

\section*{Acknowledgments}
This work was supported by the National Science Fund for Distinguished Young Scholars (No. 51925505), the General Program of National Natural Science Foundation of China (No. 52275491), the Major Program of the National Natural Science Foundation of China (No. 52090052), the National Key R\&D Program of China (No. 2022YFB3402600), and New Cornerstone Science Foundation through the XPLORER PRIZE.

\clearpage
\bibliographystyle{unsrt}

\bibliography{RO-NORM}

\bio{}
\endbio

\endbio

\end{document}